\documentclass{article}

\usepackage{microtype}
\usepackage{graphicx}
\usepackage{booktabs} 

\usepackage{hyperref}

\usepackage[accepted]{icml2025}

\usepackage{amsmath}
\usepackage{amssymb}
\usepackage{mathtools}
\usepackage{amsthm}

\usepackage[capitalize,noabbrev]{cleveref}
\usepackage{algorithm}
\usepackage{algorithmic}
\usepackage{bbding}
\usepackage{newfloat}
\usepackage{listings}
\usepackage{subfig}
\usepackage{caption}
\usepackage{minted}
\usepackage{amssymb} 
\usepackage{amsmath}
\usepackage{tikz}
\usepackage{graphicx}
\usepackage{multirow}
\usepackage{pgfplots}
\usepackage{booktabs}
\usepackage{appendix}
\usepackage{comment}
\usepackage{enumitem}
\usepackage{pifont}
\usepackage{adjustbox}
\usepackage{tikz}
\usepackage{makecell}
\usepackage{nameref}
\usepackage{prettyref}
\usepackage{csquotes}
\usepackage{fontawesome5}
\usepackage{xcolor}
\usepackage[listings,skins,breakable]{tcolorbox}

\usetikzlibrary{tikzmark}
\usetikzlibrary{fit}

\usetikzlibrary{shapes.geometric, arrows.meta, positioning, calc}
\usetikzlibrary{decorations.pathmorphing} 
\usetikzlibrary{matrix}

\theoremstyle{plain}

\theoremstyle{definition}

\theoremstyle{remark}

\usepackage[textsize=tiny]{todonotes}

\icmltitlerunning{OptiProxy-NAS: Optimization Proxy based End-to-End Neural Architecture Search}

\def\our{\texttt{OptiProxy-NAS}}

\newtcolorbox{quotebox}{
    colback=lightpurple,
    colframe=black!75,
    boxrule=0pt,
    top=5pt,
    bottom=5pt,
    left=8pt,
    right=8pt,
    arc=8pt,
    boxsep=0pt,
    toptitle=2pt,
    bottomtitle=2pt,
    fonttitle=\bfseries,
}
\definecolor{lightpurple}{rgb}{0.69, 0.61, 0.85} 
\newcommand{\pref}{\prettyref}
\newrefformat{sec}{Section~\ref{#1}}
\newrefformat{fig}{Figure~\ref{#1}}
\newrefformat{tab}{Table~\ref{#1}}
\newrefformat{eq}{Equation~\ref{#1}}
\newcommand{\checkedbox}{\text{$\rlap{$\checkmark$}\square$}} 
\definecolor{CBblue}{HTML}{0072B2}
\definecolor{CBorange}{HTML}{E69F00}
\definecolor{CBgreen}{HTML}{009E73}
\definecolor{CBred}{HTML}{D55E00}
\definecolor{CBpurple}{HTML}{CC79A7}
\definecolor{CBbrown}{HTML}{A6761D}

\begin{document}

\twocolumn[
\icmltitle{OptiProxy-NAS: Optimization Proxy based End-to-End\\ Neural Architecture Search}



\icmlsetsymbol{equal}{*}

\begin{icmlauthorlist}
\icmlauthor{Bo Lyu}{zjl}
\icmlauthor{Yu Cui}{zjl}
\icmlauthor{Tuo Shi}{zjl}
\icmlauthor{Ke Li}{sch}
\end{icmlauthorlist}

\icmlaffiliation{zjl}{Zhejiang Lab, Hangzhou, China}
\icmlaffiliation{sch}{Department of Computer Science, University of Exeter, Exeter, UK}

\icmlcorrespondingauthor{Bo Lyu}{bo.lyu@zhejianglab.org}
\icmlcorrespondingauthor{Ke Li}{k.li@exeter.ac.uk}

\icmlkeywords{Machine Learning, ICML}

\vskip 0.3in
]



\printAffiliationsAndNotice{}  

\begin{abstract}
Neural architecture search (NAS) is a hard computationally expensive optimization problem with a discrete, vast, and spiky search space. One of the key research efforts dedicated to this space focuses on accelerating NAS via certain proxy evaluations of neural architectures. Different from the prevalent predictor-based methods using surrogate models and differentiable architecture search via supernetworks, we propose an \textit{optimization proxy} to streamline the NAS as an end-to-end optimization framework, named \our. In particular, using a proxy representation, the NAS space is reformulated to be continuous, differentiable, and smooth. Thereby, any differentiable optimization method can be applied to the gradient-based search of the relaxed architecture parameters. Our comprehensive experiments on $12$ NAS tasks of $4$ search spaces across three different domains including computer vision, natural language processing, and resource-constrained NAS fully demonstrate the superior search results and efficiency. Further experiments on low-fidelity scenarios verify the flexibility.
\end{abstract}

\section{Introduction}
Neural architecture search (NAS) aims to automate the design of high-performing neural architectures for specific tasks. It has been successfully employed to conduct a \textit{de-novo} design of neural network architectures that offer better accuracy versus latency trade-offs than the best human-engineered alternatives. On the other hand, NAS has an enormously large discrete architecture search space with non-convex and isolated optimum fitness landscape, making it notoriously challenging. There have been significant efforts on {NAS sampling algorithm based on black-box optimization techniques}, including evolutionary algorithm (EA)~\textcolor{gray}{\cite{sun2020automatically,wei2022npenas}}, Bayesian optimization (BO)~\textcolor{gray}{\cite{DBLP:conf/aaai/WhiteNS21/BANANA}}, and reinforcement learning (RL)~\textcolor{gray}{\cite{zoph2017neural, zoph2018learning}}. Because the performance evaluation of each candidate architecture requires a computationally demanding deep neural networks training routine, a long-standing open problem in NAS research is how to strike a balance between the trade-off of the search efficiency and the quality of the obtained architectures. We refer the interested readers to two pieces of excellent survey papers~\textcolor{gray}{\cite{xie2021weight,DBLP:journals/corr/abs-2301-08727}} for more detailed discussions.
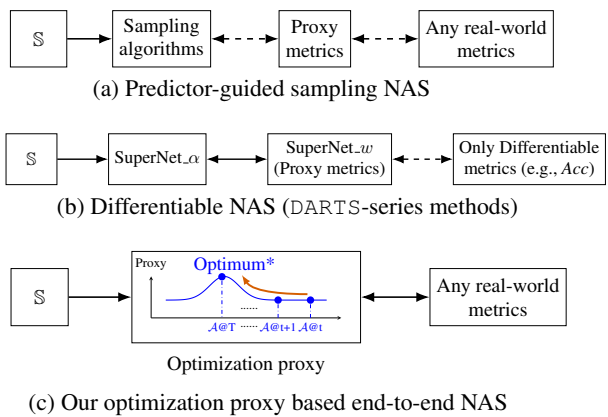
\begin{figure}[t!]
  \subfloat[{Predictor-guided sampling NAS}\label{fig:t-flow-1}]{
    \resizebox{0.46\textwidth}{!}{\usetikzlibrary{positioning}
\usetikzlibrary{calc}
\pgfplotsset{compat=1.17}

\definecolor{CBblue}{HTML}{0072B2}
\definecolor{CBorange}{HTML}{E69F00}
\definecolor{CBgreen}{HTML}{009E73}
\definecolor{CBred}{HTML}{D55E00}
\definecolor{CBpurple}{HTML}{CC79A7}
\definecolor{CBbrown}{HTML}{A6761D}

\begin{tikzpicture}[
  line/.style={draw, -Latex},
  line_d/.style={draw, Latex-Latex},
    scale=0.65
]
    \node[draw, rectangle, minimum width=1cm, minimum height=1cm] (alphaBeta) at (0,0) {$\mathbb{S}$};

    \node[draw, rectangle, minimum width=1cm, minimum height=1cm, right=0.8cm of alphaBeta, align=center] (search) {Sampling\\ algorithms};

    \draw[line, thick] (alphaBeta.east) -- (search.west);

    \node[draw, rectangle, minimum width=1cm, minimum height=1cm, right=1.2cm of search, align=center] (surr) {Proxy \\metrics};

    \draw[line_d, dashed, thick] (search.east) -- (surr.west);

    \node[draw, rectangle, minimum width=1cm, minimum height=1cm, right=1.2cm of surr, align=center] (eval) {Any real-world \\metrics};

    \draw[line_d, dashed, thick] (surr.east) -- (eval.west);

\end{tikzpicture}

}
  } \\
  \subfloat[Differentiable NAS (\texttt{DARTS}-series methods)\label{fig:t-flow-darts}]{
    \resizebox{0.5\textwidth}{!}{\usetikzlibrary{positioning}
\usetikzlibrary{calc}
\pgfplotsset{compat=1.17}

\definecolor{CBblue}{HTML}{0072B2}
\definecolor{CBorange}{HTML}{E69F00}
\definecolor{CBgreen}{HTML}{009E73}
\definecolor{CBred}{HTML}{D55E00}
\definecolor{CBpurple}{HTML}{CC79A7}
\definecolor{CBbrown}{HTML}{A6761D}

\begin{tikzpicture}[
  line/.style={draw, -Latex},
  line_d/.style={draw, Latex-Latex},
    scale=0.65
]
    \node[draw, rectangle, minimum width=1cm, minimum height=1cm] (alphaBeta) at (0,0) {$\mathbb{S}$};

    \node[draw, rectangle, minimum width=1cm, minimum height=1cm, right=1cm of alphaBeta, align=center,] (search) {SuperNet\_$\alpha$};

    \draw[line, thick] (alphaBeta.east) -- (search.west);

    \node[draw, rectangle, minimum width=1cm, minimum height=1cm, right=1.2cm of search, align=center] (surr) {SuperNet\_$w$\\(Proxy metrics)};

    \draw[line_d, thick] (search.east) -- (surr.west);

    \node[draw, rectangle, minimum width=1cm, minimum height=1cm, right=1.2cm of surr, align=center] (eval) {Only Differentiable\\ metrics (e.g., \textit{Acc})};


    \draw[line_d, dashed, thick] (surr.east) -- (eval.west);

\end{tikzpicture}

}
  } \\
  \subfloat[Our optimization proxy based end-to-end NAS\label{fig:t-flow-2}]{
    \resizebox{0.47\textwidth}{!}{\usetikzlibrary{positioning}
\usetikzlibrary{calc}
\pgfplotsset{compat=1.17}

\definecolor{CBblue}{HTML}{0072B2}
\definecolor{CBorange}{HTML}{E69F00}
\definecolor{CBgreen}{HTML}{009E73}
\definecolor{CBred}{HTML}{D55E00}
\definecolor{CBpurple}{HTML}{CC79A7}
\definecolor{CBbrown}{HTML}{A6761D}

\begin{tikzpicture}[
  line/.style={draw, -Latex},
    line_d/.style={draw, Latex-Latex},
scale=0.65
]
    \node[draw, rectangle, minimum width=1cm, minimum height=1cm] (alphaBeta) at (0,0) {$\mathbb{S}$};

    \node[draw, rectangle, minimum width=4cm, minimum height=1.6cm, right=1.1cm of alphaBeta, anchor=west] (box) {};

    \begin{axis}[
        name=plot, 
        at={(box.center)}, 
        anchor=center, 
        domain=-3:10, 
        samples=100,
        xlabel={},
        ylabel={Proxy},
        axis lines=left,
        xtick=\empty,
        ytick=\empty,
        no markers,
        width=6.7cm, 
        height=2.7cm, 
        clip=false,
        enlargelimits=false,
        xmin=-3, xmax=10, 
        ymin=0, ymax=0.7,   
        label style={font=2pt, at={(ticklabel* cs:1)}, anchor=near ticklabel}, 
        tick label style={font=\small}, 
        every axis x label/.style={
            at={(ticklabel* cs:1.05)}, anchor=west
        },
        every axis y label/.style={
            at={(ticklabel* cs:1.05)}, anchor=south
        }
    ]

    \node[anchor=north, font=\small] at (axis cs:3.7, 0.25) {......};
    \node[anchor=north, font=\small] at (axis cs:3.7, -0.05) {......};
    \addplot[blue, thick, domain=-2:9] {1/(sqrt(2*pi)) * exp(-(x-2)^2/2) + 0.30} 
    node[pos=0.35, circle, fill, inner sep=2pt] (dot1) {} 
    node[pos=0.70, circle, fill, inner sep=2pt] (dot2) {} 
    node[pos=0.9, circle, fill, inner sep=2pt] (dot3) {};
    
    \draw[dash dot, blue] (dot1) -- (dot1 |- {axis cs:0,0})  node[below, font=\small] {$\mathcal{A}$@T};
    \draw[dash dot, blue] (dot2) -- (dot2 |- {axis cs:0,0}) node[below, font=\small] {$\mathcal{A}$@t+1};
    \draw[dash dot, blue] (dot3) -- (dot3 |- {axis cs:0,0})  node[below, font=\small] {$\mathcal{A}$@t};
    
    \node[] (start) at (axis cs:8, 0.4) {};
    \node[] (end) at (axis cs:3, 0.8) {};
    \draw[-latex, thick, line width=1.5pt, color=CBred] (start) .. controls (axis cs:6.5, 0.4) and (axis cs:4, 0.42) .. (end);
    \node at (axis cs:2.7, 1.1) [anchor=north, blue, font=\Large] {Optimum*};

    \end{axis}

    \node[below=0.1cm of box] {Optimization proxy};

    \node[draw, rectangle, minimum width=1cm, minimum height=1cm, right=1.2cm of box, align=center] (eval) {Any real-world \\metrics};

    \draw[line_d, thick] (box.east) -- (eval.west);

    \draw[line, thick] (alphaBeta.east) -- (box.west);
    
\end{tikzpicture}

}
  }
  \caption{Schematic of the search framework comparison.}
  \label{fig:concept}
\end{figure}

{In recent years, NAS techniques converge to two routines.
\begin{itemize}[noitemsep,nolistsep]
    \item The first one focuses on developing proxy metrics as predictors to guide the architecture sampling while maintaining a limited evaluation cost, as illustrated in~\pref{fig:concept}(a). Representative proxies include surrogate models~\textcolor{gray}{\cite{wei2022npenas,DBLP:conf/aaai/WhiteNS21/BANANA,DBLP:conf/nips/WhiteZRLH21}}, learning curve extrapolation \textcolor{gray}{\cite{DBLP:conf/nips/YanWSH21}}, and zero-cost metrics \textcolor{gray}{\cite{DBLP:conf/cvpr/ZhouZZLYZO20,DBLP:conf/iclr/AbdelfattahMDL21}}. However, these proxies have been criticized for their unreliable modeling capability, evaluation biases, and inconsistent performance across tasks~\textcolor{gray}{\cite{colin2022adeeperlook}}. The predictor-guided sampling NAS still requires a significant number of full training iterations of architectures~\cite{DBLP:journals/corr/abs-2301-08727}. 

    \item The other one mainly uses one-shot methods to reduce the overall evaluation cost. Its basic idea is to train a supernetwork (\texttt{DARTS}-series \textcolor{gray}{\cite{liu2019darts,DBLP:conf/iclr/XieZLL19,DBLP:conf/cvpr/YeL00FO22,DBLP:conf/iclr/CaiZH19}}) or hypernetwork (SMASH \textcolor{gray}{\cite{DBLP:conf/iclr/BrockLRW18}}) and then use that to provide a quick individual (subnetwork) evaluation. As illustrated {in~\pref{fig:concept}(b)}, \texttt{DARTS}-series methods~\textcolor{gray}{\cite{liu2019darts,DBLP:conf/iclr/XieZLL19,DBLP:conf/cvpr/YeL00FO22}} construct a supernetwork to enable differentiable optimization over relaxed architectural variables, where the loss function serves as a proxy for \textit{accuracy}. However, this formulation inherently makes it challenging to incorporate non-differentiable metrics, such as \textit{latency} and \textit{power}, along with high memory consumption in practical search \textcolor{gray}{\cite{xie2021weight,DBLP:journals/corr/abs-2301-08727}}.
\end{itemize}

\vspace{-.5em}
Given these limitations, we contend to reach the trade-off between full-pipeline resource cost and resulting practically deployable architectures, not only optimized for \textit{accuracy} but also account for customizable objectives such as resource constraints, runtime latency, and other deployment-specific metrics. Unlike previous efforts primarily centered on constructing proxies for costly real-world evaluation metrics, such as the \textit{architecture-accuracy} mapping, we come up with a different hypothesis as follows.}

\begin{quotebox}
    \noindent
    \faLightbulb \, \textit{{If the NAS space is represented by a proxy space that is continuous, differentiable, and smooth—particularly one that provides a well-structured representation of high-performance regions—the search process can become more targeted and directly guided by real-world metrics, even enabling gradient-based optimization.}}
\end{quotebox}

Building upon this hypothesis, this paper proposes a \underline{Opti}mization-\underline{Proxy} based end-to-end \underline{NAS} framework, dubbed as \our.
\begin{itemize}[noitemsep,nolistsep]
    \item The basic idea of \our\ is to synthesize the prediction and search into an \textit{optimization proxy}, which formulates NAS as an end-to-end optimization problem in a differentiable way.

    \item \our\ does not require a proxy representation of the entire space. Instead, it progressively models the NAS landscape by starting from a coarse estimation and refining it towards the precise promising regions with the search progression.

    \item \our\ is flexible as it enables a direct incorporation of any real-world NAS metrics, e.g., latency constraint and low-fidelity evaluation metrics.
\end{itemize}

Generally speaking, \our\ is featured in the following three advantages.
\begin{itemize}[noitemsep,nolistsep]
    \item As illustrated in~\pref{fig:concept}(c), our end-to-end differentiable optimization enables the gradient information flow among the discrete search space, the \textit{optimization proxy} and any real-world metrics. Comparatively, as predictor-guided sampling NAS shown in~\pref{fig:concept}(a), the samplers (RL, EA, or BO) operate in discrete space and are isolated with proxy metrics, preventing direct gradient flow. Thus, we ensure superior sampling efficiency compared to them.

    \item Unlike \texttt{DARTS}-series methods, our \textit{optimization proxy} reduces the disparity between the discrete architecture space and real-world metrics. It avoids costly supernetworks, and expands the search beyond merely differentiable objectives, such as \textit{accuracy}. {This makes it more suitable for not only the deployment platform-aware NAS scenarios (typically involving both {differentiable} and {non-differentiable} search objectives) but also the resource-sensitive scenarios in the search process itself.}

    \item {Our \textit{optimization proxy} relies merely on the labeled DAG representation of architecture, a standard encodings scheme in NAS. This ensures a wide applicability of \our\ across various search spaces.}
    
\end{itemize}

The performance of \our\ is validated in a comprehensive experiment covering diverse scenarios, including $6$ single-objective spaces and $6$ multi-objective cases, a large space with up to $10^{53}$ architectures and $12$ NAS tasks ranging from CV, NLP, resource-constrained (under hardware-aware metrics of deployment devices), to multi-fidelity.
The results fully demonstrate the effectiveness and superiority of \our\ against {15} state-of-the-art (SOTA) peer algorithms. Specifically, $\blacktriangleright$ on NAS-Bench-201 \textcolor{gray}{\cite{DBLP:conf/iclr/dong2020bench}}, we consistently achieve the optimum with nearly {$100\%$} less query cost across all tasks; $\blacktriangleright$ on NAS-Bench-301 \textcolor{gray}{\cite{DBLP:journals/corr/siems2020bench}} and NAS-Bench-NLP \textcolor{gray}{\cite{klyuchnikov2022bench}}, we achieve significantly superior results and {$400\%$} efficiency improvement; $\blacktriangleright$ on HW-NAS-Bench \textcolor{gray}{\cite{DBLP:conf/iclr/LiYFZZYY0HL21}}, under {$12$} latency-constrained settings, across {$6$} hardware devices, our results are consistently the best comparing to the published best results with the same queries, and outperforms SOTA methods in all {$12$} device settings even with {$50\%$} fewer queries;
$\blacktriangleright$ we also experimentally show the compatibility and flexibility of \our\ under low-fidelity evaluation settings, and with $4$ types of proxy models. In particular, \our\ adds merely $20$ seconds and $10$MB of additional memory, costing only $0.037\%$ runtime and $0.25\%$ memory of the total NAS pipeline. Our code will be released at \url{https://github.com/blyucs/OptiProxy-NAS}.


\section{Proposed Method}

\subsection{Problem Formulation}
Conventionally, NAS is formulated as a bi-level optimization problem within the discrete space:
\begin{equation}
\begin{gathered}
\max_{\mathcal{A} \in \mathbb{S}} \quad \operatorname{Eval}(\langle\mathcal{A}, w^{*}(\mathcal{A})\rangle; \mathbb{D}_\mathtt{val}) \\
\text{s.t.} \quad w^{*}(\mathcal{A})=\arg \min_{w} \mathcal{L}(w(\mathcal{A}); \mathbb{D}_\mathtt{train}),
\end{gathered}
\label{eq:A_star}
\end{equation}
where $\mathcal{A}$ stands for the architecture and $\mathbb{S}$ represents the architecture space, the lower-level optimization focuses on the training of the candidate architectures on $\mathbb{D}_\mathtt{train}$ and $\mathcal{L}$ is a loss function, while the upper-level optimization aims to search for a neural architecture that maximizes the evaluation accuracy on $\mathbb{D}_\mathtt{val}$, denoted as $f$. 
Following predictor-based NAS works \textcolor{gray}{\cite{wen2020neural,wei2022npenas,dudziak2020brp,DBLP:conf/aaai/WhiteNS21/BANANA}}, given the model $\hat{f}$ (with parameters $\theta$) that provides the approximate representation of the landscape of $f$, the NAS problem can be approximated as:
\begin{equation}
\begin{array}{ll} \label{eq:optimizer-darts}
\max \limits_{\mathcal{A} \in \mathbb{S}} & \hat{f}\left(\mathcal{A}, \theta^{*}\right),
\end{array}
\end{equation}
where $\theta^*$ denotes the optimal parameters of $f$. 
Given that we focus solely on the differentiable models, which are proven to be effective as metric predictors, to reach an ideal model ($\theta^*$) is infeasible due to the astronomical combinatorial space. Instead, we may approximately solve it by empirical risk minimization, formulated as:
\begin{equation}
\label{eq:heuristic}
\begin{aligned}
& \max \limits_{\mathcal{A} \in \mathbb{S}} \hat{f}\left(\mathcal{A}, \theta^{*}\right) \\
\text{ s.t. } \quad & \theta^{*} = \underset{\theta}{\mathrm{argmin}} \sum_{\mathcal{A} \in \mathcal{D}} \mathcal{L}(\hat{f}_\theta(\mathcal{A}), f(\mathcal{A})), \\
\end{aligned}
\end{equation}
where $\mathcal{L}$ is the empirical loss function, and $\mathcal{D} = \{(\mathcal{A}_i, f(\mathcal{A}_i))\}_{i=1}^B$ is the sampled dataset on $\mathbb{S}$. Specifically, the upper-level optimization is the maximization of $\hat{f}$ over $\mathcal{A}$, while the lower-level optimization includes fitting the model on $\mathcal{D}$.
Back to NAS space, the network architectures can be parameterized by operations features (candidate operation choices) $\mathcal{A}_{o}$ and the topological structure $\mathcal{A}_{t}$. 
Even $\hat{f}$ is capable of fitting the landscape to be continuous, $\mathcal{A}_{o}$ and $\mathcal{A}_{t}$ are both discrete, thus the space remains discrete. Borrowed from \textcolor{gray}{\cite{DBLP:conf/iclr/XieZLL19, DBLP:conf/iclr/CaiZH19, liu2019darts}}, we have $\mathcal{A}_o$ and $\mathcal{A}_t$ follow a probability distribution $\alpha$ and $\beta$, respectively, allowing the architecture variable to be relaxed. We denote this continuous space spanning by $\alpha$ and $\beta$ as $\mathcal{X}$, the proxy space.
Thus, the optimization problem is reformulated as:
\begin{equation}
\label{eq:final_formulation}
\begin{aligned}
& \max \limits_{(\alpha,\beta) \in \mathcal{X}} \hat{f}(s(\alpha, \beta), \theta^{*}), \\
\text{s.t.} \quad & \theta^{*} = \underset{\theta}{\mathrm{argmin}} \sum_{\mathcal{A} \in \mathcal{D}} \mathcal{L},(\hat{f}_\theta(\mathcal{A}), f(\mathcal{A})) \\
&  \mathcal{D} = \{(\mathcal{A}_b, f(\mathcal{A}_b))\}_{b=1}^B,\\
&  \mathcal{A}_b \sim s(\alpha,\beta),\\
\end{aligned}
\end{equation}
where ${s}$ is the architecture sampling function. The lower-level optimization targets \(\theta\) (model fitting), depending on the data \(\mathcal{D}\) that collected by $s$ following \(\alpha,\beta\). The upper-level optimization focuses on \(\alpha,\beta\). Although this formulation keeps the form of bi-level optimization, both the upper- and lower-level optimizations target parts of the $\hat{f}(\cdot)$, namely the input variables and weights, it can be solved end-to-end based on gradients. Borrowed from \textcolor{gray}{\cite{DBLP:journals/corr/abs-2112-13469}}, we treat this search paradigm as ``{Optimization Proxy}'' for NAS.
\subsection{Gradient-based Search in Optimization Proxy}
Our design of the search commences with the sampling function $s(\alpha,\beta)$. First, we have $\mathcal{A}_o$ follows a categorical distribution with the probability vector set $\alpha \in \mathbb{R}^{N \times M}$, $N$ represents feature nodes number and $M$ is the candidate operation number, for node $i$:
\begin{equation}
\label{eq:categorial}
\mathcal{A}_{o}^{i} \sim \text{Cat}(\textit{softmax}(\alpha^i)),
\end{equation}
where $\textit{softmax}(\alpha^i)=(p^0, p^1, \ldots, p^{M-1})$ represents the sampling probabilities of each of the \(M\) candidate operations, \( 0\leq p^j \leq 1 \) and \(\sum_{j=0}^{M-1} p^j = 1\).\\
Further, we have $\mathcal{A}_{t}$ distributed according to a Bernoulli distribution with \(\beta \in \mathbb{R}^{N \times N}\), for a specific connection:
\begin{equation}
\label{eq:bernoulli}
\mathcal{A}_{t}^{(h,k)} \sim \text{Bernoulli}(\sigma(\beta^{(h,k)})),
\end{equation}
where \(\sigma (\beta^{(h,k)}) \) denotes the occurrence probability of the connection between node $h$ and $k$, within the range of $[0,1]$. 
\begin{algorithm}[t!]
\caption{Optimization Proxy based Architecture Search}
\label{alg:acq_function}
\begin{algorithmic}[1]
\STATE \textbf{Input}: $(\alpha$, $\beta)$, $\hat{f}_{\theta^*}$;
\STATE \textbf{Parameters}: \textit{oEpochs}, \textit{tEpochs}, \textit{searchEpochs}, Initial Gumbel Temperature $\tau$, learning rate $\eta_1$, $\eta_2$;
\STATE \textbf{Output}: The promising architecture batch $\{ \mathcal{A}_b \}_{b=1}^Q$;
\STATE  Fix the proxy model weights ${\theta^*}$;
\FOR{$e = 1$ to \textit{searchEpochs}}
    \STATE $\hat{s}_o$, $\hat{s}_t$ $\gets$ $\hat{s}(\alpha)$, $\hat{s}({\beta})$; \hfill  \textcolor{purple}{$\triangleright$ \texttt{Proxy sampling}}
    \STATE out $\gets$ $\hat{f}_{\theta^*}(\hat{s}_o, \hat{s}_t$);   \hfill \textcolor{purple}{$\triangleright$ \texttt{Forward}}
    \STATE loss $\gets$ out();  \hfill \textcolor{purple}{$\triangleright$ \texttt{Output maximization}}
    \STATE loss.backward() to achieve $\nabla_\alpha \psi(\alpha, \beta)$, $\nabla_\beta \psi(\alpha, \beta)$; \\ \hfill \textcolor{purple}{$\triangleright$ \texttt{Gradients calculated}}
    \STATE Alternatively call:  \hfill \textcolor{purple}{$\triangleright$ \texttt{Gradients applied}}\\
     $\quad \alpha_{t+1} = \alpha_t + \eta_1 \nabla_\alpha \psi(\alpha, \beta, \theta^*)$; \\
     $\quad \beta_{t+1} = \beta_t + \eta_2 \nabla_\beta \psi(\alpha, \beta, \theta^*)$; \\
        for \textit{oEpochs} and \textit{tEpochs} intervals, respectively.  \\ 
    \STATE Decay the $\tau$;
\ENDFOR
\STATE $\{ \mathcal{A}_b \}_{b=1}^Q$ $\gets$ $s(\alpha^*,\beta^*)$ by Eq. \eqref{eq:categorial},\eqref{eq:bernoulli}.  \hfill \textcolor{purple}{$\triangleright$ \texttt{Sampling}}
\end{algorithmic}
\end{algorithm}
Following Eq. \eqref{eq:final_formulation}, the architecture search is to maximize the objective function $\psi= \hat{f}({s}(\alpha,\beta), \theta^*)$.
The gradients of the objective function w.r.t \(\alpha\) and \(\beta\) are as follow:
\begin{equation}\label{eq:grad0}
\begin{aligned}
\nabla_\alpha \psi = \frac{\partial \hat{f}}{\partial {s}} \cdot \frac{\partial {s}}{\partial \alpha} \text{,} \quad  
\nabla_\beta \psi = \frac{\partial \hat{f}}{\partial {s}} \cdot \frac{\partial {s}}{\partial \beta},
\end{aligned}
\end{equation}
where \({\partial \hat{f}}/{\partial {s}}\) is the gradient of the $\hat{f}$ output w.r.t input feature, which follows $\theta^*$. \textit{As the sampling function $s$ is in discrete space, \textcolor{black}{Eq. \eqref{eq:grad0}} is not calculable, we need a proxy sampling function $\hat{s}$ that satisfies the fundamental conditions: $\blacktriangleright$ Differentiable; $\blacktriangleright$ Approximate discrete sampling $s$ with unbiased relaxation.} 
Hence, the gradients may be approximated:
\begin{equation}\label{eq:surgrad0}
\begin{aligned}
\nabla_\alpha \psi \approx \frac{\partial \hat{f}}{\partial \hat{s}} \cdot \frac{\partial \hat{s}}{\partial \alpha} \text{,} \quad 
\nabla_\beta \psi \approx \frac{\partial \hat{f}}{\partial \hat{s}} \cdot \frac{\partial \hat{s}}{\partial \beta},
\end{aligned}
\end{equation}
We treat these as the {proxy gradients} of the objective function w.r.t $(\alpha, \beta)$.

We use the re-parameterize trick to realize $\hat{s}$. Concretely, operation feature $\hat{s}_o$ are drawn from the {Gumbel softmax} distribution \textcolor{gray}{\cite{DBLP:conf/iclr/JangGP17}}, for node $i$:
\begin{equation}\label{eq:forward_ops}
\begin{aligned}
\hat{s}_{o}^{(i,j)} & = \frac{\exp \left(\left( \alpha^{(i,j)}+g^{(i,j)}\right) / \tau\right)}{\sum_{m=0}^{M-1} \exp \left(\left( \alpha^{(i,m)}+g^{(i,m)}\right) / \tau\right)},
\end{aligned}
\end{equation} 
where $0\leq i < N-1$, $0\leq j < M-1$, and $g^{(i,j)}$ are {i.i.d} samples drawn from Gumbel{$(0, 1)$}, \(\tau\) is the temperature within $[0, \infty)$. The $\alpha^{(i,j)}$ stands for the unnormalized-logit probability value of the candidate operation $j$ at node $i$. Further, for topological structure samples $\hat{s}_t$:
\begin{equation}\label{eq:forward_adj}
\begin{aligned}
\hat{s}_{t}^{(h,k)} & = \sigma \left(\left(\beta^{(h,k)}+g^{(h,k)}\right) / \tau\right),
\end{aligned}
\end{equation}
where \( g^{(h,k)} \) are i.i.d samples drawn from \(\text{Gumbel}(0,1)\) for \(0 \leq h, k < N-1 \), and $\sigma$ is the sigmoid function, which maps input to range $(0,1)$. The $\beta^{(h,k)}$ represents the log-odds value of the implicit probability variable $\gamma^{(h,k)}$, that is, $\beta^{(h,k)}=\text{log}({\gamma^{(h,k)}}/{(1-\gamma^{(h,k)})})$, $\gamma^{(h,k)} \in [0,1]$.

We present the optimization proxy based search pseudocode as~\pref{alg:acq_function}. {Given the ideal proxy model $\hat{f}_{\theta^*}$, for one search epoch, the proxy model iteratively takes the output} of \textcolor{purple}{\textit{Proxy sampling}} function ($\hat{s}$) as input, it then undergoes the \textcolor{purple}{\textit{Forward}} pass, and directly takes model output as the maximization item for \textcolor{purple}{\textit{Gradients calculation}}. Subsequently, $(\alpha,\beta)$ are updated by \textcolor{purple}{\textit{Gradients applied}}, which are alternatively carried out over the predefined intervals, respectively. Finally, after optimization of specific epochs, the promising architecture batch \textcolor{purple}{\textit{Sampling}} based on currently optimized $(\alpha^*,\beta^*)$ are taken as search algorithm output.
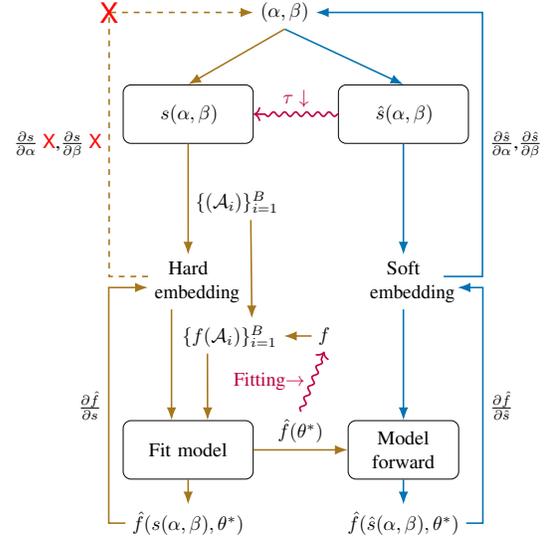
\begin{figure}[!ht]
  \centering
    \resizebox{0.95\linewidth}{!}{ 



\definecolor{CBblue}{HTML}{0072B2}
\definecolor{CBorange}{HTML}{E69F00}
\definecolor{CBgreen}{HTML}{009E73}
\definecolor{CBred}{HTML}{D55E00}
\definecolor{CBpurple}{HTML}{CC79A7}
\definecolor{CBbrown}{HTML}{A6761D}

\tikzset{
  block/.style={rectangle, draw, text width=6em, text centered, rounded corners, minimum height=3em},
  line/.style={draw=CBblue, thick, -Latex},
  line2/.style={draw=CBbrown, thick, -Latex},
  text_large/.style={scale=1}, 
}

\begin{tikzpicture}[scale=0.1]

  \node[text_large] (alpha_beta) at (0, 0) {$(\alpha, \beta)$};
  
  \node[block, below right=1cm and 0.4cm of alpha_beta] (gumbel) {$\hat{s}(\alpha,\beta)$ }; 

  \node[text_large, below=2cm of gumbel, text width=1.2cm, align=center] (e2) {Soft \\embedding};

  \node[block, below=2cm of e2, text width=5em] (model_forward) {Model forward}; %
  
  \node[text_large, below=0.45cm of model_forward] (gradient) {$\hat{f}(\hat{s}(\alpha,\beta), \theta^*)$};

 \node[block, below left=1cm and 0cm of alpha_beta] (multinomial) {$s(\alpha,\beta)$};

  
  \node[text_large, below =2cm of multinomial, text width=1.2cm, align=center] (e1) {Hard \\embedding};

  \node[text_large, below right =0.2cm and 1.5cm of e1] (f) {$f$};

  \node[text_large, left=0.5cm of f] (D) {$\{f(\mathcal{A}_i)\}_{i=1}^B$};

  \node[block, below=2cm of e1] (fit_model) {Fit model};

  \node[text_large, below=0.45cm of fit_model] (f_hat) {$\hat{f}(s(\alpha,\beta), \theta^*)$};

  \path[line] (alpha_beta.south) -- node[above] {} (gumbel.north);
  
  \path[line] (gumbel.south) -- node[left] {} (e2.north);

    
      \draw[->, decorate, line width=0.3mm, decoration={snake, amplitude=.4mm, segment length=2mm, post length=1mm}, purple]
        (gumbel.west) -- (multinomial.east) node[midway, above, color=purple] {\textcolor{purple}{$\tau \downarrow$}};



        
  \path[line2] (multinomial.south) -- node[] {} ++(0, -1) coordinate(midpoint) -- node[right] (ori_D) {$\{(\mathcal{A}_i)\}_{i=1}^B$} (e1.north);



  \path[line2] (f.west) -- node[right] {} (D.east);

\path[line2] ([xshift=-4cm]D.south) -- ([xshift=3.5cm]fit_model.north);
\path[line2] ([xshift=-3cm]e1.south) -- ([xshift=-3cm]fit_model.north);

  \path[line] (model_forward.south) --  node[below] {} (gradient.north);

\path[line2] (alpha_beta.south) -- node[above] {} (multinomial.north);

  \path[line2] (fit_model.south) -- node[above] {} (f_hat.north);
  
  \path[line2] (fit_model.east) -- node[above] (f_theta) {$\hat{f}(\theta^*)$} (model_forward.west);
    \draw[->, decorate, line width=0.3mm, decoration={snake, amplitude=.4mm, segment length=2mm, post length=1mm},purple]
        (f_theta.north) -- ([xshift=0cm]f.south) node[midway, left] {\textcolor{purple}{Fitting$\rightarrow$}};

  \path[line2] ([xshift=2.5cm]ori_D.south) -- node[above] {} ([xshift=4cm]D.north);
        
  \path[line] (e2.south) -- node[above] {} (model_forward.north);

\path[line] (gradient.east) -- ++(3,0) |- ([yshift=-1cm,xshift=2.5cm]e2.east) 
    node[right, pos=0.25, align=left] {$\frac{\partial \hat{f}}{\partial \hat{s}}$};

\path[line2] (f_hat.west) -- ++(-3,0) |- ([yshift=-1cm]e1.west) 
    node[left, pos=0.25, align=left] {$\frac{\partial \hat{f}}{\partial {s}}$};

\path[line] ([yshift=1cm]e2.east) -- ++(7,0) |- ([yshift=0cm,xshift=0cm]alpha_beta.east) 
    node[right, pos=0.25, align=left] {$\frac{\partial \hat{s}}{\partial {\alpha}}$,$\frac{\partial \hat{s}}{\partial {\beta}}$};

\path[line2, dashed] ([yshift=1cm]e1.west) -- ++(-7,0) |- ([yshift=0cm]alpha_beta.west) 
    node[left, pos=0.25, align=left] {$\frac{\partial s}{\partial {\alpha}}$ \textcolor{red}{\textsf{X}},$\frac{\partial s}{\partial {\beta}}$ \textcolor{red}{\textsf{X}}}
    node[pos=0.5, red, font=\Large] {\textsf{X}}; 

\end{tikzpicture}

  }
  \caption{Illustration of the differentiable search strategy. The discrete sampling \(s\) renders the path non-differentiable. With the introduction of proxy sampling \(\hat{s}\), it makes the propagation differentiable.} 
  \label{fig:surgrad}
\end{figure}
\begin{algorithm}[t!]
\caption{\our\ framework}
\begin{algorithmic}[1]
\STATE \textbf{Input}: $({\alpha},{\beta}$), Proxy model $\hat{f}_{\theta}$;
\STATE \textbf{Parameters}: Evaluation function $f$;  Maximum evaluation number $C$; Sampling number $Q$ and the evaluation number $B$ of each step; 
\STATE \textbf{Output}: The optimal $\mathcal{A}^*$;
\STATE $\mathcal{D} \gets $ {Randomly sample several architectures;}
\WHILE{$n \leq C$}
    \STATE Train the model $\hat{f}_\theta$ using $\mathcal{D}$;  \hfill \textcolor{purple}{$\triangleright$ \texttt{Fitting}}
    \STATE Search $\{\mathcal{A}_b \}_{b=1}^Q$ by {\color{black}Algorithm. \ref{alg:acq_function}$(\alpha,\beta)$}; \hfill \textcolor{purple}{$\triangleright$ \texttt{Search}}
    \STATE Select $\{\mathcal{A}_b\}_{b=1}^B$ from $\{\mathcal{A}_b\}_{b=1}^Q$ by $\hat{f}_{\theta^*}$;\\\hfill \textcolor{purple}{$\triangleright$ \texttt{Selection}}
    \STATE Evaluate (query) the selected batch; \\ \hfill \textcolor{purple}{$\triangleright$ \texttt{Evaluation}}
    \STATE $\mathcal{D} \gets \mathcal{D} \cup \{(\mathcal{A}_b,f(\mathcal{A}_b))\}_{b=1}^B$; \\ \hfill \textcolor{purple}{$\triangleright$ \texttt{Dataset update}}
    \STATE $n \gets n + B$; 
\ENDWHILE
\STATE $\mathcal{A}^* \gets$ The historical best architecture in $\mathcal{D}$.
\end{algorithmic}
\label{alg:framework}
\end{algorithm}
\begin{figure}[!ht]
  \centering
    \resizebox{0.9\linewidth}{!}{ 
  {\input{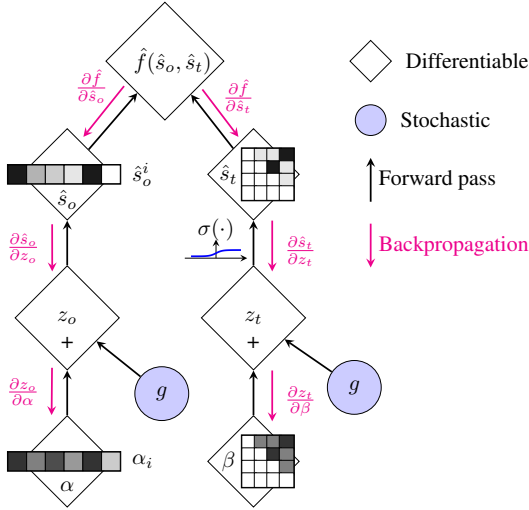}}
  }
  \caption{Gradient calculation in the computation graph.} 
  \label{fig:gumbel}
\end{figure}
\subsection{Design Intentions and Analysis}
As aforementioned, we explicitly propose two sets of proximal relationships:
$\hat{f} \Longleftrightarrow f$, $\hat{s} \Longleftrightarrow s$, which derive four sets of proximal relationships: $\partial{\hat{f}}/\partial{s} \Longleftrightarrow \partial{f}/\partial{s}$, $\partial{\hat{f}}/\partial{\hat{s}} \Longleftrightarrow \partial{\hat{f}}/\partial{s}$, $\partial{\hat{s}}/\partial{\alpha} \Longleftrightarrow \partial{s}/\partial{\alpha}$, $\partial{\hat{s}}/\partial{\beta} \Longleftrightarrow \partial{s}/\partial{\beta}$. We present the design intentions in \textcolor{black}{\pref{fig:surgrad}}. The discrete sampling and model fitting (\(\hat{f}\) approaching to \(f\)) implicitly establish the original gradients, as \textcolor{black}{Eq. \eqref{eq:grad0}}, but renders the propagation path non-differentiable. With the design of $\hat{s}$, we approximately achieve the full-path differentiable propagation. The analysis is summarized as follows:
\begin{itemize}
\item {Unbias}: For the operations feature variable, in \textcolor{gray}{\cite{DBLP:conf/iclr/MaddisonMT17}}, it has been proved that:
\begin{equation}\label{eq:lambda_alpha}
\begin{aligned}
p\left( \lim_{\tau \to 0} \hat{s}_{o}^{(i,j)} = 1 \right) = \textit{softmax}(\alpha^{i})^{j},
\end{aligned}
\end{equation}
which is the categorical variables in \textcolor{black}{Eq. \eqref{eq:categorial}}. In terms of topological structure variables, it also have been proved that:
\begin{equation}\label{eq:lambda_beta}
\begin{aligned}
p( \lim_{\tau \to 0} \hat{s}_{t}^{(h,k)} & = 1 )  = \gamma^{(h,k)},
\end{aligned}
\end{equation}
which is the Bernoulli distribution variables in \textcolor{black}{Eq. \eqref{eq:bernoulli}}. We present these detailed proofs in \textcolor{black}{Appendix \ref{sec:proof}}. Upon this, as $\tau \to 0$, then $ \hat{s} \to s $, \text{and then:}
\begin{equation}\label{eq:approach}
\begin{aligned}
\frac{\partial \hat{f}}{\partial \hat{s}} \cdot \frac{\partial \hat{s}}{\partial \alpha} \to \frac{\partial \hat{f}}{\partial {s}} \cdot \frac{\partial {s}}{\partial \alpha}\text{,} \quad \frac{\partial \hat{f}}{\partial \hat{s}} \cdot \frac{\partial \hat{s}}{\partial \beta} \to \frac{\partial \hat{f}}{\partial {s}} \cdot \frac{\partial {s}}{\partial \beta},
\end{aligned}
\end{equation}
which makes the relaxation unbiased and enables the {proxy gradients} to approximate the original gradients.
\item {Differentiable}: As in \textcolor{black}{\pref{fig:gumbel}}, the re-parameterization allows gradients to flow from $\hat{f}$ output to both $\alpha$ and $\beta$, to obtain the proxy gradients.
\item {Unconstrained}: The optimization of continuous variables proceeds without constraint, that is, $\alpha \in \mathbb{R}^{N \times M}$, \(\beta \in \mathbb{R}^{N \times N}\).
\item {Exploration-exploitation dilemma:} The $\mathcal{A}_o$ sampling by \textcolor{black}{Eq.\eqref{eq:categorial}} and $\mathcal{A}_t$ sampling by \textcolor{black}{Eq.\eqref{eq:bernoulli}} introduce both exploration and exploitation. In the re-parameterization, the Gumbel noise variable $g$ introduces randomness to $(\alpha,\beta)$, facilitating exploration. During the early stages of search, a large $\tau$ encourages exploration, while in the later stages, as $\tau \to 0$, it facilitates exploitation.
\end{itemize}
\begin{table*}[!h]
\caption{Comparison of results on NAS-Bench-201 cross different datasets. We search on \texttt{val} accuracy and report the mean$\pm$standard deviation result over $10$ runs of different random seeds and the query number from the benchmark.}
\label{tab:nas_results}
\resizebox{0.99\textwidth}{!}{%
\setlength{\tabcolsep}{3mm}
\begin{tabular}{@{}lcccccp{2.5cm}c@{}}
\toprule
\multirow{2}{*}{\textbf{Method}} & \multicolumn{2}{c}{\textbf{CIFAR-10}} & \multicolumn{2}{c}{\textbf{CIFAR-100}} & \multicolumn{2}{c}{\textbf{ImageNet16-120}} & \multirow{2}{*}{\textbf{Queries}} \\ 
\cmidrule(lr){2-3} \cmidrule(lr){4-5} \cmidrule(lr){6-7}
 & Val (\%) & Test (\%) & Val (\%) & Test (\%) & Val (\%) & Test (\%) &  \\ \midrule
\textbf{Optimum*} & $\mathbf{91.61}$ & $\mathbf{94.37}$ & $\mathbf{73.49}$ & $\mathbf{73.51}$ & $\mathbf{46.77}$ & $\mathbf{47.31}$ & $-$ \\ \midrule
SGNAS \textcolor{gray}{\cite{DBLP:conf/cvpr/HuangC21}} & $90.18$ & $93.53$ & $70.28$ & $70.31$ & $44.65$ & $44.98$ & $-$ \\
BANANAS$^{\dagger}$\textcolor{gray}{\cite{DBLP:conf/aaai/WhiteNS21/BANANA}} & $91.56$ & $94.30$ & $\mathbf{73.49*}$ & $73.50$ & $46.65$ & $46.51$ & $192$ \\
Bayesian Opt. $^{\dagger}$ \textcolor{gray}{\cite{DBLP:conf/icml/SnoekRSKSSPPA15}} & $91.54$ & $94.22$ & $73.26$ & $73.22$ & $46.43$ & $46.40$ & $192$ \\
Random Search $^{\dagger}$ \textcolor{gray}{\cite{DBLP:conf/uai/LiT19}} & $91.12$ & $93.89$ & $72.08$ & $72.07$ & $45.97$ & $45.98$ & $192$ \\
GA-NAS \textcolor{gray}{\cite{DBLP:conf/ijcai/RezaeiHNSMLLJ21}} & $-$ & $94.34$ & $-$ & $73.28$ & $-$ & $\mathbf{46.80^*}^\ddagger$ & $444$ \\
MetaD2A \textcolor{gray}{\cite{DBLP:conf/iclr/LeeHH21}} & $94.37$ & $94.38$ & $73.34$ & $73.34$ & $-$ & $-$ & $500$ \\ 
$\beta$-DARTS \textcolor{gray}{\cite{DBLP:conf/cvpr/YeL00FO22}} & $91.55$ & $94.36$ & $\mathbf{73.49^*}$ & $\mathbf{73.51^*}$ & $46.37$ & $46.34$ & $-$ \\
TNAS \textcolor{gray}{\cite{DBLP:conf/iclr/ShalaEHG23}} & $-$ & $\mathbf{94.37^*}$ & $-$ & $\mathbf{73.51^*}$ & $-$ & $-$ & $-$ \\ \midrule
DiffusionNAG \textcolor{gray}{\cite{DBLP:journals/corr/abs-2305-16943}} & $-$ & $\mathbf{94.37^*}$ & $-$ & $73.51$ & $-$ & $-$ & $-$ \\ \midrule \rule{0pt}{2.5ex}  
DiNAS \textcolor{gray}{\cite{DBLP:journals/corr/abs-2403-06020}}  & \tikzmark{start7} $\mathbf{91.61^*}$ & $\mathbf{94.37^*}$ & $\mathbf{73.49^*}$ & $\mathbf{73.51^*}$ \tikzmark{end7} & $46.66$ & $45.41$ & $192$ \\ \midrule \rule{0pt}{2.5ex}  
\multirow{3}{*}{AG-Net \textcolor{gray}{\cite{DBLP:journals/corr/abs-2203-08734}}} & $91.41$ & $94.16$ & $73.14$ & $73.15$ & $46.42$ & $46.43$ \tikzmark{end1} & $100$ \\ \rule{0pt}{2.5ex}  
 & $91.60$ & $\mathbf{94.37^*}$ & $\mathbf{73.49^*}$ & $\mathbf{73.51^*}$ & \tikzmark{start3} $46.64$ & $46.43$ \tikzmark{end3} &  $192$ \\  \rule{0pt}{2.5ex}  
 & \tikzmark{start2} $\mathbf{91.61^*}$  & $\mathbf{94.37^*}$ & $\mathbf{73.49^*}$ & $\mathbf{73.51^*}$ \tikzmark{end2} &  \tikzmark{start5} $46.73$ & $46.42$ \tikzmark{end5} & $400$  \\ \midrule \rule{0pt}{2.5ex}  
\multirow{3}{*}{\textbf{\our\ (Ours)}}  & \tikzmark{start1} $\mathbf{91.61}^*$ & $\mathbf{94.37^*}$ & $\mathbf{73.49^*}$ & $\mathbf{73.51^*}$ \tikzmark{end1} &  \tikzmark{start4} $\mathbf{46.64}\pm0.12$ & $45.98^\ddagger$  \tikzmark{end4} & $\mathbf{100}$ \\ \rule{0pt}{2.5ex} 
& $-$  & $-$ & $-$ & $-$ & $\mathbf{46.69}\pm0.1$ & ${45.79}^\ddagger$ & $\mathbf{192}$ \\ \rule{0pt}{2.5ex} 
& $-$ & $-$ & $-$ & $-$ & \tikzmark{start6} $\mathbf{46.77^*}$ & ${45.47}^\ddagger$ \tikzmark{end6} & $\mathbf{280^*}$ \\   \bottomrule 
\end{tabular}
\begin{tikzpicture}[overlay, remember picture]
    \coordinate (start1) at ($(pic cs:start1) + (0, -0.3em)$);
    \coordinate (end1) at ($(pic cs:end1) + (0, 1em)$);
    \coordinate (start2) at ($(pic cs:start2) + (0, -0.3em)$);
    \coordinate (end2) at ($(pic cs:end2) + (0, 1em)$);

    \coordinate (start7) at ($(pic cs:start7) + (0, -0.3em)$);
    \coordinate (end7) at ($(pic cs:end7) + (0, 1em)$);

    \coordinate (start3) at ($(pic cs:start3) + (0, -0.3em)$);
    \coordinate (end3) at ($(pic cs:end3) + (0, 1em)$);
    \coordinate (start4) at ($(pic cs:start4) + (0, -0.3em)$);
    \coordinate (end4) at ($(pic cs:end4) + (0, 1em)$);

    \coordinate (start5) at ($(pic cs:start5) + (0, -0.3em)$);
    \coordinate (end5) at ($(pic cs:end5) + (0, 1em)$);
    \coordinate (start6) at ($(pic cs:start6) + (0, -0.3em)$);
    \coordinate (end6) at ($(pic cs:end6) + (0, 1em)$);
    
    \draw[dashed, thick, purple] (start1) rectangle (end1);
    \draw[dashed, thick, purple] (start2) rectangle (end2);
    
    \draw[dashed, thick, red] (start3) rectangle (end3);
    \draw[dashed, thick, red] (start4) rectangle (end4);

    \draw[dashed, thick, blue] (start5) rectangle (end5);
    \draw[dashed, thick, blue] (start6) rectangle (end6);

    \draw[dashed, thick, purple] (start7) rectangle (end7);

    \draw[->, thick, diamond, purple] ([shift={(0,0.5em)}]start2) .. controls +(-0.5,0) and +(-0.5,0) .. node[pos=0.8, left, , xshift=1.5em] {\parbox{3em}{$4\times$ \\queries}} ([shift={(0,0.3em)}]start1);

    \draw[->, thick, diamond, purple] ([shift={(0,0.5em)}]start7) .. controls +(-0.5,0) and +(-0.5,0) .. node[pos=0.5, left, , xshift=1.5em] {\parbox{3em}{$\approx 2\times$ \\queries}} ([shift={(0,0.9em)}]start1);

    \draw[->, thick, diamond, red] ([shift={(0,-0.5em)}]end4) .. controls +(0.4,0) and +(0.4,0) .. node[pos=0.7, right, xshift=-1.7em] {\parbox{4em}{$\approx 50\%$ \\fewer queries}} ([shift={(0,-0.5em)}]end3);
    \draw[->, thick, diamond, blue] ([shift={(0,-0.5em)}]end6) .. controls +(0.4,0) and +(0.4,0) .. node[pos=0.3, right, xshift=-1.8em] {\parbox{5em}{Reach ``$*$''\\even with\\$70\%$ queries}} ([shift={(0,-0.5em)}]end5);
    
\end{tikzpicture}
}\\
\begin{flushleft}
\footnotesize{$^\dagger$: Results taken from \textcolor{gray}{\cite{DBLP:journals/corr/abs-2203-08734}}.}
\footnotesize{$^\ddagger$: Following other works, we search on \texttt{val} accuracy to derive the optimal architecture and report its corresponding \texttt{test} accuracy. This metric (\texttt{test}) should not serve as an indicator of search performance.}
\end{flushleft}
\end{table*}
\subsection{Architecture Search Framework}
Here we apply the sequential model based optimization (SMBO) as the main search framework. The pseudocode of our search framework is given in~\pref{alg:framework}, the search begins with a randomly initialized dataset $\mathcal{D}$, iteratively undergoing \textcolor{purple}{\textit{{Model Fitting}}, {\textit{Search}}, {\textit{Selection}}, {\textit{Evaluation}}}, and \textcolor{purple}{\textit{Dataset Update}}, as one search step. {As $\mathcal{D}$ is iteratively updated with newly evaluated architecture that undergoes optimization proxy based \textit{search} and proxy model based \textit{selection}, the fitting of proxy model gradually achieves the better representation of the high-performance architecture regions.} The iterative process continues for specific query cost constraints, to achieve the best historically evaluated architecture as the NAS result. The schematic of \our\ is presented in the \textcolor{black}{Appendix \ref{sec:frameworks}}.
Typically, an effective exploration strategy to avoid traps in local optima is crucial amidst a vast and potentially noisy space landscape.
We have $\boldsymbol{\alpha} = [\alpha_1, \alpha_2, \ldots, \alpha_K]$ and $\boldsymbol{\beta} = [\beta_1, \beta_2, \ldots, \beta_K]$ denote the batches of parameters, where each $\alpha_i$ and $\beta_i$ are initialized using Latin Hypercube Sampling (LHS). In one search step, considering $Q$ candidate architectures, we sample \( {Q}/{K} \) architectures from each \( (\alpha_i,\beta_i) \) group. This strategy attempts to enhance the sampling diversity and prevents the clustering in specific regions, which facilitates a systematic exploration. Fortunately, different groups of \((\alpha_i,\beta_i)\) share the query metrics and the proxy model, making it almost zero additional cost.
\subsection{Proxy Model and Encoding Scheme}
In theory, any differentiable model, not limited to the type and structure, combined with the corresponding architecture encoding scheme can work in \our\ framework. In this paper, we adopt a preliminary graph structure encoding scheme that treats candidate operations as nodes and feature maps as edges. The model takes the original one-hot feature matrix and one-hot adjacency matrix as inputs \textcolor{gray}{\cite{DBLP:conf/nips/WhiteNNS20encoding}}. A detailed explanation of encoding schemes for the search spaces is provided in Appendix \ref{sec:space_encoding}.

\begin{table*}[!h]
\caption{Comparison results on NB101, NB301 and NBNLP. We search on \texttt{val} accuracy and report the mean$\pm$standard deviation results over $10$ runs with different random seeds.}
\label{tab:combined_nasbench_results}
\resizebox{0.93\textwidth}{!}{
\setlength{\tabcolsep}{0.5mm}
\begin{tabular}{@{}lccccp{2.5cm}cc@{}}
\toprule
\multirow{2}{*}{\textbf{Method}} & \multicolumn{3}{c}{\textbf{NAS-Bench-101}} & \multicolumn{2}{c}{\textbf{NAS-Bench-301}} & \multicolumn{2}{c}{\textbf{NAS-Bench-NLP}} \\
\cmidrule(lr){2-4} \cmidrule(lr){5-6} \cmidrule(lr){7-8}
 & Val (\%) & Test (\%) & Queries & Val (\%) & Queries & Val (\%) & Queries \\
\midrule
\textbf{Optimum*} & $\mathbf{95.06}$ & $\mathbf{94.32}$ & $-$ & $-$ & $-$ & $-$ & $-$ \\ \midrule
Arch2vec+RL \textcolor{gray}{\cite{DBLP:conf/nips/YanZAZ020}} & $-$ & $94.10$ & $400$ & $-$ & $-$ & $-$ & $-$ \\
Arch2vec+BO \textcolor{gray}{\cite{DBLP:conf/nips/YanZAZ020}} & $-$ & $94.05$ & $400$ & $-$ & $-$ & $-$ & $-$ \\
NAO $^{\dagger}$ \textcolor{gray}{\cite{DBLP:conf/nips/LuoTQCL18}}  & $94.66$ & $93.49$ & $192$ & $-$ & $-$ & $-$ & $-$ \\
Random Search $^{\dagger}$ \textcolor{gray}{\cite{DBLP:conf/uai/LiT19}}  & $94.31$ & $93.61$ & $192$ & $94.31$ & $195$ & $95.64$ & $304$ \\
Regularized Evolution $^{\dagger}$ \textcolor{gray}{\cite{real2019regularized}} & $94.47$ & $93.89$ & $192$ & $94.75$ & $192$ & $95.66$ & $304$ \\
BANANAS $^{\dagger}$ \textcolor{gray}{\cite{DBLP:conf/aaai/WhiteNS21/BANANA}} & $94.73$ & $94.09$ & $192$ & $94.47$ & $192$ & $95.68$ & $304$ \\
Local Search $^{\dagger}$ \textcolor{gray}{\cite{DBLP:conf/uai/WhiteNS21}} & $94.57$ & $93.97$ & $192$ & $-$ & $-$ & $-$ & $-$ \\
WeakNAS \textcolor{gray}{\cite{DBLP:conf/nips/WuDCCLYWLCY21}} & $-$ & $94.18$ & $200$ & $-$ & $-$ & $-$ & $-$ \\
DiNAS \textcolor{gray}{\cite{DBLP:journals/corr/abs-2403-06020}} & $94.98$ & $\mathbf{94.27}$ & $150$ & $94.92$ & $100$ & $96.06$ & $304$ \tikzmark{e4} \\ \rule{0pt}{2.5ex}
AG-Net \textcolor{gray}{\cite{DBLP:journals/corr/abs-2203-08734}} & $94.90$ & $94.18$ & $192$ & \tikzmark{s2} $94.79$ & $192$ \tikzmark{e2}  & \tikzmark{s4} $95.95$ & $304$ \\\midrule \rule{0pt}{2.5ex}
\multirow{2}{*}{\textbf{\our\ (Ours)}} & $\mathbf{94.98}\pm0.16$ & $94.18^{\ddagger}$ & $\mathbf{142}$ & $\mathbf{94.97}\pm0.05$ & $\mathbf{100}$ & $\mathbf{96.18}\pm0.20$ & $\mathbf{150}$ \\ \rule{0pt}{2.5ex}  
 & $-$ & $-$ & $-$ & \tikzmark{s1} $\mathbf{94.81}\pm0.07$ & $\mathbf{\ \ 50}$ \tikzmark{e1} & \tikzmark{s3} $\mathbf{96.08}\pm0.19$ & $\mathbf{75}$ \tikzmark{e3} \\
\bottomrule
\end{tabular}

\begin{tikzpicture}[overlay, remember picture]
    \coordinate (s1) at ($(pic cs:s1) + (0, -0.4em)$);
    \coordinate (e1) at ($(pic cs:e1) + (0.5em, 1em)$);
    \coordinate (s2) at ($(pic cs:s2) + (0, -0.4em)$);
    \coordinate (e2) at ($(pic cs:e2) + (0, 1em)$);
    
    \coordinate (s3) at ($(pic cs:s3) + (0, -0.2em)$);
    \coordinate (e3) at ($(pic cs:e3) + (0.5em, 1em)$);
    \coordinate (s4) at ($(pic cs:s4) + (0, -0.2em)$);
    \coordinate (e4) at ($(pic cs:e4) + (0, 1em)$);

    
    \draw[dashed, thick, purple] (s1) rectangle (e1);
    \draw[dashed, thick, purple] (s2) rectangle (e2);
    
    \draw[dashed, thick, blue] (s3) rectangle (e3);
    \draw[dashed, thick, blue] (s4) rectangle (e4);


    \draw[->, thick, diamond, purple] ([shift={(0,-0.5em)}]e1) .. controls +(0.4,0) and +(0.4,0) .. node[pos=0.3, right, xshift=-1.2em] {\parbox{4em}{$\sim4\times$\\efficiency}} ([shift={(0,-0.6em)}]e2);

    \draw[->, thick, diamond, blue] ([shift={(0,-0.5em)}]e3) .. controls +(0.4,0) and +(0.4,0) .. node[pos=0.4, right, xshift=-1.5em] {\parbox{4em}{$\sim4\times$\\efficiency}} ([shift={(0,-1em)}]e4);

    
\end{tikzpicture}

}\\
\begin{flushleft}
\footnotesize{$^\dagger$: Results taken from \textcolor{gray}{\cite{DBLP:journals/corr/abs-2203-08734}}.} 
\footnotesize{$^\ddagger$: The global optimal $95.06$ (\texttt{val} accuracy) arch's corresponding \texttt{test} accuracy is $94.23$, we reach the optimum* over $8/10$ of all seeds. We search on \texttt{val} accuracy, as stated, \texttt{test} metric does not represent the efficacy.}
\end{flushleft}
\end{table*}


\vspace{-.5em}
\section{Experimental Results}
\label{sec:results}
\begin{table}[ht]
\caption{Comparison results on HW-NAS-Bench. We report the mean results over $10$ runs with different random seeds.}
\label{tab:hw_nas_results}
\resizebox{0.47\textwidth}{!}{%
\setlength{\tabcolsep}{0.5mm}
\begin{tabular}{@{}lcp{2.3cm}p{2.5cm}ccc@{}}
\toprule
\multirow{2}{*}{\centering Device} & \multirow{2}{*}{\centering Constraint} & \multicolumn{2}{c}{\textbf{\our}} & AG-Net  & DiNAS & \multirow{2}{*}{Optimum*} \\
\cmidrule(lr){3-4} \cmidrule(lr){5-5} \cmidrule(lr){6-6} 
                         &            & 200 (queries)      & 100 (\textcolor{purple}{$50\%$ fewer})     &   200 $^{\dagger}$    &    200 $^{\dagger}$    &          \\
\midrule
\multirow{3}{*}{EdgeGPU} & $2$            & $\mathbf{40.54}\pm0.19$ & \underline{${40.12}\pm0.41$} & $39.70$      &  $39.44$      & $40.60$        \\ 
                         & $4$            & $\mathbf{44.56}\pm0.23$ & \underline{${44.15}\pm0.63$} & $42.80$      &  $43.91$      & $44.80$        \\ 
                         & $6$            & $\mathbf{46.02}\pm0.23$ & \underline{${45.65}\pm0.61$} & $45.30$      &  $45.03$      & $46.40$        \\ \midrule
\multirow{3}{*}{Raspi4}  & $2$            & $\mathbf{34.98}\pm0.22$ & \underline{${34.91}\pm0.25$} & $34.60$      &  $34.67$      & $35.47$        \\ 
                         & $4$            & $\mathbf{43.36}\pm0.29$ & \underline{${43.19}\pm0.32$} & $42.00$      &  $43.25$      & $43.63$        \\ 
                         & $6$            & $\mathbf{44.75}\pm0.39$ & \underline{${44.69}\pm0.28$} & $44.00$      &  $44.72$      & $45.17$        \\ \midrule
EdgeTPU                  & $1$            & $\mathbf{46.55}\pm0.11$ & \underline{${46.52}\pm0.12$} & $46.40$      &  $45.31$     & $46.77$       \\ \midrule
\multirow{3}{*}{Pixel3}  & $2$            & $\mathbf{41.10}\pm0.48$ & \underline{${41.04}\pm0.49$} & $40.90$      &  $40.01$     & $41.30$       \\ 
                         & $4$            & $\mathbf{45.86}\pm0.32$ & \underline{${45.46}\pm0.48$} & $45.30$      &  $44.74$      & $45.97$        \\ 
                         & $6$            & $\mathbf{46.19}\pm0.29$ & \underline{${45.97}\pm0.29$} & $45.70$      &  $45.95$      & $46.48$        \\ \midrule
Eyeris                   & $1$            & $\mathbf{45.03}\pm0.48$ & \underline{${44.83}\pm0.10$} & $44.50$      &  $44.67$      & $45.16$        \\ \midrule
FPGA                     & $1$            & $\mathbf{44.03}\pm0.19$ & \underline{${43.85}\pm0.08$} & $43.30$      &  $\mathbf{44.53}^\ddagger$     & $\mathbf{44.37}^\ddagger$        \\ 
\bottomrule
\end{tabular}
}\\
\footnotesize{$^\dagger$: Results taken from \textcolor{gray}{\cite{DBLP:journals/corr/abs-2403-06020}}. $^\ddagger$: According to our tabular query, the optimum under constraint $1$ms of FPGA is $44.37\%$.}
\end{table}
To validate the effectiveness of \our, we follow the NAS best practice (see \textcolor{black}{Appendix \ref{sec:prac}}) to set a fair empirical study. We choose $15$ SOTA algorithms focusing on sample efficiency under the mainstream search strategies for comparison (see detailed experiment setup in \textcolor{black}{Appendix \ref{sec:exp_details}}).
To demonstrate the compatibility, we conduct preliminary experiments with MLP, LSTM and GAT.
In particular, we report the comparison results based on the graph convolutional network (GCN) proposed by~\textcolor{gray}{\cite{wen2020neural}} (see details in Appendix \ref{sec:surrogatemodel}).
\subsection{Tabular and Surrogate NAS Benchmarks}
We first report the comparison results on four tabular and surrogate benchmarks.\\
\textbf{NAS-Bench-201.} 
\pref{tab:nas_results} presents the comparison results of \our\ against the other $11$ peer algorithms on the NAS-Bench-201 benchmark. Specifically, for the CIFAR-10 and CIFAR-100 datasets, \our\ obtains the reported optimum architecture with only $100$ queries. Although \texttt{DiNAS} and \texttt{AG-Net} also find the optimum, they cost $2\times$ and $4\times$ more queries, respectively. From a visual diagnostic analysis of the space and its exploration paths in \pref{fig:visual} of CIFAR-10 in Appendix \ref{sec:visual}, we find that sampling based on $(\alpha, \beta)$ is rapidly guided to the high-performance region, significantly enhancing the overall sampling efficiency. We believe such improvement is benefited from the proxy representation of the NAS space and gradient-based optimizer. For ImageNet$16$-$120$, the model scale setting in NAS-Bench-201 results in a relatively aggravated task difficulty and distribution complexity, making the space landscape appear noisier, which poses challenges for proxy representation and optimization. Theoretically, it is relatively hard to achieve a trade-off between query cost and performance on ImageNet16-120. Nevertheless, \our\ achieves the optimum with $280$ queries, while no other peer algorithms have reported achieving this, e.g., \texttt{AG-Net} fails even with $400$ queries. Compared to \texttt{AG-Net}, which costs at least $192$ queries to obtain the near-optimal results ($46.64$), \our\ is more efficient as it costs $\approx 50\%$ fewer queries to reach the same results.
\begin{figure*}[!ht]
\centering
\subfloat[]{
\includegraphics[width=5.6cm]{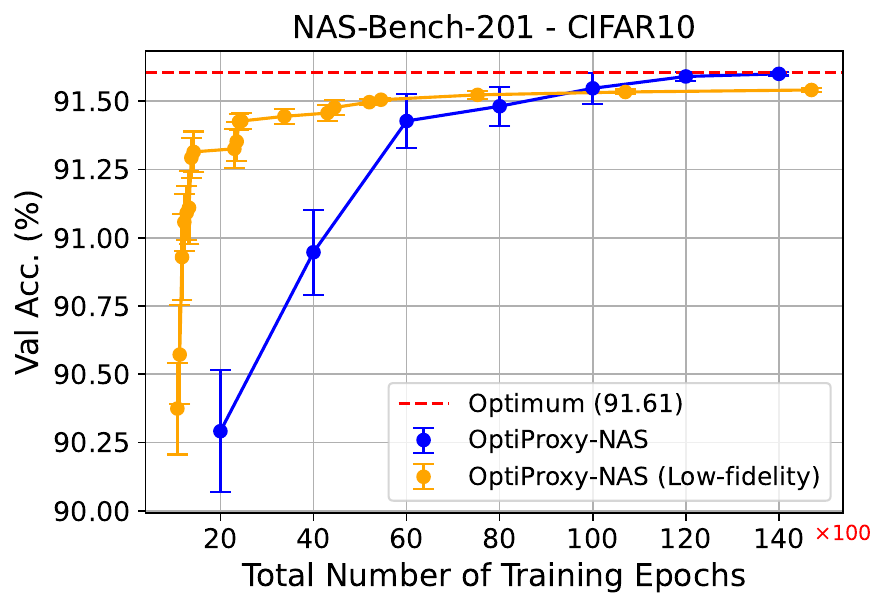}
}
\subfloat[]{
\includegraphics[width=5.4cm]{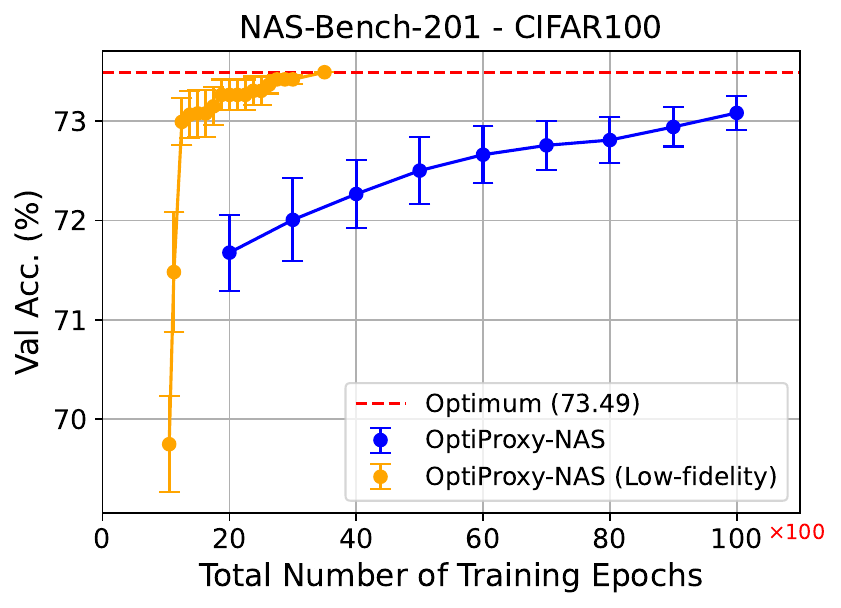}
}
\subfloat[]{
\includegraphics[width=5.6cm]{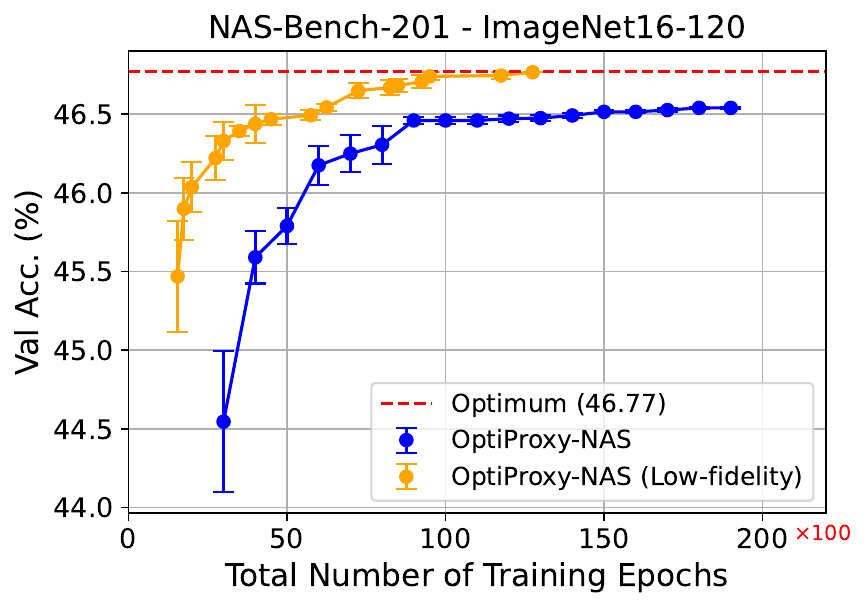}
}
\caption{Comparison results of low-fidelity (5 epochs) search and full training (200 epochs) search (\our).}
\label{fig:multi_fidelity_1}
\end{figure*}\\
\textbf{NAS-Bench-101.} The search space of the NAS-Bench-101 benchmark is more challenging due to the existence of \lq isomorphic graphs\rq\ and \lq validity constraints\rq, which introduce substantial optimization errors and perturbations when learning the topological structures and operation node features. Due to the page limit, detailed discussion upon the limitation and complexity is in Appendix \ref{sec:appendix_101}. The results in~\pref{tab:combined_nasbench_results} also demonstrate the superiority of \our, which achieves the best of $94.98\pm0.16$ with $142$ queries.\\
\textbf{NAS-Bench-301 and NAS-Bench-NLP.}
For the surrogate benchmarks, NAS-Bench-301 (\texttt{DARTS} space) and NAS-Bench-NLP (RNN space). As ~\pref{tab:combined_nasbench_results} shows, the performance of \our\ is not only the best but also it costs much less queries (at least $4\times$ more efficient than the other $10$ peer algorithms). This result is anticipated. Because these spaces are represented by surrogate models, they are relatively smoother and thus facilitating our proxy representation as well as the gradient-based end-to-end search.\\
Additionally, to better demonstrate the effectiveness and stability throughout the search, we compare the performance of various methods in the best ever-sampled architecture over the query times. As in~\pref{fig:query_efficient} of Appendix \ref{sec:query}, \our\ shows superior results compared with the baseline methods, especially on NAS-Bench-301 and NAS-Bench-NLP. 
\subsection{Hardware-aware NAS Benchmark}
To demonstrate the flexibility of \our, we further experiment on the resource-constrained NAS tasks of HW-NAS-Bench. {To this end, we integrate an additional regression head into the proxy model to create a proxy representation of the search space spanning with architectures and latency metrics.} In this unified optimization proxy, the proxy representations for these two cross-domain metrics share the architecture features but model the different landscape functions individually.
Further, we use a weighted aggregation of the accuracy and latency as an evaluation metric. In particular, the weights are learnable based on the latency statistics of samples to provide latency constraint management. As shown in~\pref{tab:hw_nas_results}, across $6$ of hardware devices (NN accelerators, edge and embedded devices), under $12$ latency constraint settings, the performance of \our\ constantly outperforms the published best results. {Significantly, we achieve better results than \texttt{AG-Net} in all $12$ device latency settings, even with 50\% fewer queries.}

\subsection{Low-fidelity Experiment}
To assess the compatibility and flexibility, we conduct another experiment under a low-fidelity setting, where only $5$ training epochs are allocated. We search for promising architectures batch using the low-fidelity metric and subsequently evaluate them with the high-fidelity metric to identify the best one. As shown in {\pref{fig:multi_fidelity_1}}, compared with the full training accuracy based search, the low-fidelity accuracy based search can achieve satisfying results within a limited number of total training epochs. For CIFAR-10, with an increase in the total epochs, the full training search gradually performs better due to the unbiased evaluation. For CIFAR-100 and ImageNet16-120, the low-fidelity search is consistently better than the full training search. We present the comparison results of total training epochs between low-fidelity search and full training search on NAS-Bench-201 in \pref{tab:low-fidelity}. To reach the same mean \texttt{val} accuracy, the low-fidelity search demonstrates a significant efficiency advantage.
\begin{table}[t!]
\centering
\caption{Comparison of total training epochs on NB201.} 
\label{tab:low-fidelity}
\resizebox{0.43\textwidth}{!}{%
\setlength{\tabcolsep}{1mm}
\begin{tabular}{@{}lccc@{}}
\toprule
Setting                & C10 ($91.46$) & C100 ($73.49$) & ImageNet ($46.65$) \\ \midrule
\our\                & $6000$              & $20000$          & $20000$        \\
\our\ (Low-fidelity)           & $\mathbf{3200}$     & $\mathbf{3500}$  & $\mathbf{7250}$         \\ \bottomrule
\end{tabular}
}
\end{table}
\subsection{Ablation Study}
\begin{table}[t!]
\centering
\caption{Ablation study results. The query cost settings are $100/142/100/150$ for NB201/NB101/NB301/NBNLP.} 
\label{tab:ablation}
\setlength{\tabcolsep}{0.5mm}
\resizebox{0.44\textwidth}{!}{
\begin{tabular}{@{}p{1.2cm}p{1.2cm}cccccccc@{}}
\toprule
\multirow{2}{*}{\makecell{Method\\/Setting}}               & \multirow{2}{*}{\makecell{No. of \\ $(\alpha_i,\beta_i)$}}   & \multirow{2}{*}{\texttt{Selection}}  & \multicolumn{3}{c}{NB201} & \multirow{2}{*}{NB101} & \multirow{2}{*}{NB301} & \multirow{2}{*}{NBNLP} \\
\cline{4-6}
                    &                            &          & C10 & C100 & ImageNet & & & \\\midrule
\textbf{*}   & $-$   & $-$        & $91.61$   & $73.49$   & $46.77$ & $95.06$ & - & - \\\midrule
\texttt{RS}     & $-$      & $-$        & $91.12$   & $72.08$   & $45.97$ & $94.31$ & $94.31$ & $95.64$ \\
\texttt{S0}  & $-$         & \XSolidBrush & $91.08$ & $69.56$  & $45.65$ & $94.55$ & $94.33$ & $95.80$ \\
\texttt{S1}  & $-$         & \Checkmark & $91.51$   & $72.91$   & $46.30$ & $94.59$ & $94.66$ & $95.90$ \\
\texttt{S2}  & single        & \XSolidBrush & $91.54$ & $72.30$   & $45.93$ & $94.69$ & $94.64$ & $95.79$ \\
\texttt{S3}  & multi        & \XSolidBrush & $91.49$ & $72.52$  & $46.39$ & $94.77$ & $94.71$ & $95.83$ \\
\texttt{S4}  & single       & \Checkmark & $91.57$   & $73.42$   & $46.41$ & $94.92$ & $94.82$ & $95.94$ \\
\texttt{S5}  & multi        & \Checkmark & $\mathbf{91.61^*}$  & $\mathbf{73.49^*}$ & $\mathbf{46.64}$ & $\mathbf{94.98}$ & $\mathbf{94.97}$ & $\mathbf{96.18}$ \\
\bottomrule
\end{tabular}}
\end{table}
We conduct an ablation study regarding three design components of \our, i.e., $(\alpha,\beta)$-based sampling, multi-$(\alpha,\beta)$ parallel exploration, and {p}roxy {m}odel based \texttt{selection}. From the results in~\pref{tab:ablation} (mean over $10$-seeds runs), we confirm that all components are of importance. Specifically, $(\alpha,\beta)$-based sampling is effective in resulting high-performance architectures. {This results demonstrates that the \texttt{selection} can consolidate the reliability of our optimization proxy based search strategy and end-to-end search framework. In addition, we also notice that the multi-$(\alpha,\beta)$ parallel exploration strategy can improve the overall exploration, leading to better performance.}

\subsection{Resource and Time Cost}
\label{sec:resource-time-cost}
The additional cost of optimization proxy mainly comes from the training and backpropagation of the proxy model, which is highly compact (e.g., a two-layers GCN, $139.7$K Parameters and $1.3$MFLOPs), see details in Appendix \ref{sec:surrogatemodel}. The model's complexity is $\mathcal{O}(N\times M\times K)+\mathcal{O}(N^2\times K)+\mathcal{O}(N\times K^2)$, where $N$ is the nodes number ($<$ dozens), $M$ is the operation numbers ($<$ dozens), and $K$ is the hidden neurons numbers, which does not scale with architecture or space. Experimentally, individually training $100$ NB201-C10 architecture needs over $15$ GPU-hours and $4$GB of RAM. Our framework adds merely $20$s and $10$MB of additional memory, costing only $0.037\%$ of runtime and $0.25\%$ of memory.


\vspace{-.5em}
\section{Related Works}
The individually sampling NAS \textcolor{gray}{\cite{zoph2017neural, sun2020automatically}} explore architectures within a discrete domain and generally rely on individual evaluation, which raises concerns about efficiency. Target this, predictor-guided sampling NAS employ proxy metrics \textcolor{gray}{\cite{DBLP:conf/nips/WhiteNNS20encoding,DBLP:conf/icml/WistubaP20,DBLP:conf/nips/YanWSH21}} to provide quick evaluation \textcolor{gray}{\cite{dudziak2020brp,DBLP:conf/aaai/WhiteNS21/BANANA,wei2022npenas}}. 
In recent years, zero-cost metrics enable the NAS without requiring exhaustive training \textcolor{gray}{\cite{DBLP:conf/icml/MellorTSC21,DBLP:conf/iclr/AbdelfattahMDL21}}, but they may also bias the search \textcolor{gray}{\cite{colin2022adeeperlook}}. 
In another way, the \texttt{DARTS}-series methods \textcolor{gray}{\cite{liu2019darts, chen2019progressive,DBLP:conf/iclr/XieZLL19}} relax the architecture space, by which to construct the supernetwork, then ensuring differentiable end-to-end search and weight-sharing based proxy evaluation. 
Inspired by generative models, some works reformulate NAS process as a generative paradigm, as seen in \texttt{GA-NAS}~\textcolor{gray}{\cite{DBLP:conf/ijcai/RezaeiHNSMLLJ21}} and \texttt{AG-Net}~\textcolor{gray}{\citep{DBLP:journals/corr/abs-2203-08734}}.
We provide additional details of related works in Appendix \ref{sec:search_frame}.


\vspace{-.6em}
\section{Conclusion}
\label{sec:conclusion}

This paper introduces a novel optimization proxy that redefines NAS as a differentiable optimization problem by employing a proxy representation of the search space, differentiable models, and relaxed architecture variables. Building on this foundation, we propose a differentiable search strategy that inherently balances exploration and exploitation. Furthermore, we introduce an end-to-end differentiable NAS framework (\our), streamlining the entire NAS process.
Extensive experiments across $7$ public benchmarks demonstrated it is effective, efficient, and flexible with real-world NAS metrics including platform-agnostic metrics, resource constraints and platform-specific metrics. In essence, \our\ is a general framework that can accommodate other search algorithms. We have experimentally used EA as the local search within it. For clarity, we opt to introduce the purely gradient-based framework in this paper. We believe it paves a new avenue for NAS, and we hope it will serve as a foundation that revolves with and be complementary to recent research developments in multi-fidelity evaluation, learning curve extrapolation, weight-sharing evaluation, and zero-cost metrics. We use parallelization of $(\alpha, \beta)$ to promotes diversity and tackle the isolated points. The $(\alpha_i, \beta_i)$ groups are optimized simultaneously, sharing metrics and optimization proxy. This also indicates the potential of parallelizing \our\ and we will explore this in future works.

\section*{Impact Statement}

This paper presents work whose goal is to advance the field of 
Machine Learning. There are many potential societal consequences 
of our work, none which we feel must be specifically highlighted here.


\bibliography{aaai25}
\bibliographystyle{icml2025}

\clearpage
\newpage
\appendix
\section{Appendix}
\setcounter{table}{0}
\setcounter{figure}{0}
\setcounter{equation}{0}
\subsection{The NAS Best Practices Checklist}
\label{sec:prac}
\subsubsection{Best practices for releasing code}

\textit{For all experiments you report, check if you released:}

\begin{itemize}[label=\checkedbox]
    \item \textit{Code for the training pipeline used to evaluate the final architectures.} \textcolor{gray}{[N/A]}, for fair comparison, this work regards the popular benchmarks: NAS-Bench-101, NAS-Bench-201, NAS-Bench-301, NAS-Bench-NLP, HW-NAS-Bench. The training pipelines for evaluation are released in their original repositories.
    \item \textit{Code for the search space.} \textcolor{gray}{[N/A]}.
    \item \textit{The hyperparameters used for the final evaluation pipeline, as well as random seeds} \textcolor{gray}{[N/A]}.
    \item \textit{Code for your NAS method.} \textcolor{blue}{[Yes]}, we will release our NAS framework code.
    \item \textit{Hyperparameters for your NAS method, as well as random seeds.} \textcolor{blue}{[Yes]}, we will release our hyperparameter settings, as well as the random seeds.
\end{itemize}

\textit{Note that the easiest way to satisfy the first three of these is to use existing NAS benchmarks, rather than changing them or introducing new ones.} 
\textcolor{blue}{[Yes]}, we fully agree with and comply with this practice item, to structure this work.

\subsubsection{Best practices for comparing NAS methods}

\begin{itemize}[label=\checkedbox]
    \item \textit{For all NAS methods you compare, did you use exactly the same NAS benchmark, including the same dataset (with the same training-test split), search space and code for training the architectures and hyperparameters for that code?} \textcolor{blue}{[Yes]}, this work regards the popular benchmarks: NAS-Bench-101, NAS-Bench-201, NAS-Bench-301, NAS-Bench-NLP, HW-NAS-Bench. The dataset, training-test split, search space, code for training and hyperparameters are exactly the same for fair comparison.
    \item \textit{Did you control for confounding factors (different hardware, versions of DL libraries, different runtimes for the different methods)?} \textcolor{blue}{[Yes]}, we use the same environment setting for all benchmark experiments, but running on GPU/CPU, or combined to implement the search pipeline, there exists a subtle bias on the final results. 
    \item \textit{Did you run ablation studies?} \textcolor{blue}{[Yes]}, we ran a thorough ablation study.
    \item \textit{Did you use the same evaluation protocol for the methods being compared?} \textcolor{blue}{[Yes]}.
    \item \textit{Did you compare performance over time?} \textcolor{gray}{[N/A]}, we run experiments on $7$ benchmarks for fair comparison and to compare performance over query numbers. Note that the number of queries is almost perfectly correlated with runtime.
    \item \textit{Did you compare to random search?} \textcolor{blue}{[Yes]}.
    \item \textit{Did you perform multiple runs of your experiments and report seeds?} \textcolor{blue}{[Yes]}, We ran 10 trials across all benchmarks and datasets, reporting the mean value, also with standard derivation.
    \item \textit{Did you use tabular or surrogate benchmarks for in-depth evaluations?}  \textcolor{blue}{[Yes]}, both.
\end{itemize}

\subsubsection{Best practices for reporting important details}

\begin{itemize}[label=\checkedbox]
    \item \textit{Did you report how you tuned hyperparameters, and what time and resources this required?} \textcolor{blue}{[Yes]}, in the Appendix\textcolor{blue}{~\pref{sec:hypertuning}}.
    \item \textit{Did you report the time for the entire end-to-end NAS method (rather than, e.g., only for the search phase)?} \textcolor{gray}{[N/A]}, we run experiments on $7$ benchmarks for fair comparison and to compare performance over query numbers. We do not target the evaluation acceleration strategies but have demonstrated promising results in low-fidelity experiments. Note that the number of queries is almost perfectly correlated with runtime.
    \item \textit{Did you report all the details of your experimental setup?} \textcolor{blue}{[Yes]}, we reported all details of our experimental setup in Appendix \textcolor{blue}{~\pref{sec:setup}}.
\end{itemize}

\subsection{Related works}
\label{sec:search_frame}
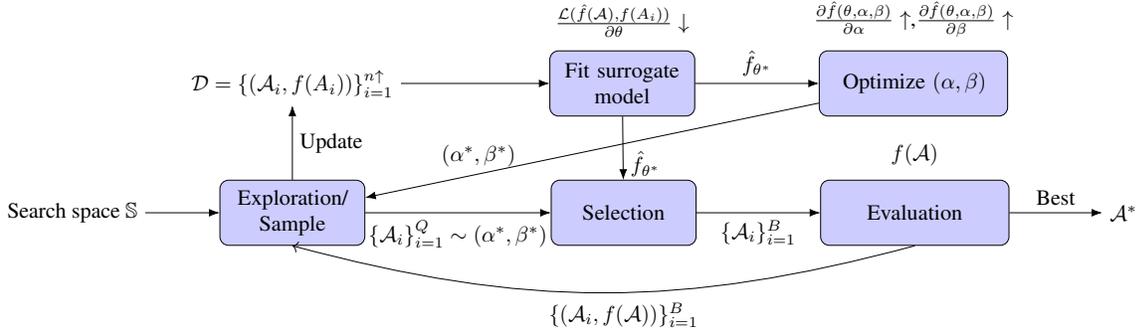
\begin{figure*}[!ht]
  \centering
      \resizebox{0.9\linewidth}{!}{ 





\begin{tikzpicture}[
  block/.style={rectangle, draw, text width=6em, fill=blue!20, text centered, rounded corners, minimum height=3em},
  line/.style={draw, -Latex},
]

\node[] (config) {Search space $\mathbb{S}$};
\node[block, right=1.2cm of config] (policy) {Exploration/ Sample};
\node[block, right=3cm of policy] (sel) {Selection};
\node[block, right=2cm of sel, text width=8em] (eval) {Evaluation};
\node[above=0.1cm of eval] {$f(\mathcal{A})$};
\node[above=1.2cm of policy, minimum height=2em] (data) {$\mathcal{D}=\{ (\mathcal{A}_i, f(A_i)) \}_{i=1}^{n \uparrow}$};
\node[block, right=2.4cm of data] (model) {Fit surrogate model};
\node[above=0.1cm of model] {$\frac{\mathcal{L}(\hat{f}(\mathcal{A}),f(A_i))}{\partial{\theta}} \downarrow$};
\node[block, right=2cm of model, text width=8em] (function) {Optimize $(\alpha, \beta)$};
\node[above=0.1cm of function] {$\frac{\partial{\hat{f}(\theta, \alpha, \beta)}}{\partial{\alpha}} \uparrow$,$\frac{\partial{\hat{f}(\theta, \alpha, \beta)}}{\partial{\beta}} \uparrow$};
\node[below=0.8cm of sel] {$\{ (\mathcal{A}_i, f(\mathcal{A})) \}_{i=1}^B$};
\node[right=1.5cm of eval] (best) {$\mathcal{A}^*$};

\draw[line] (config.east) -- (policy.west);
\draw[line] (policy.east) -- (sel.west) node[midway, below]{$\{ \mathcal{A}_i \}_{i=1}^Q \sim (\alpha^*,\beta^*)$};
\draw[line] (sel.east) -- (eval.west) node[midway, below]{$\{\mathcal{A}_i\}_{i=1}^B$};
\draw[line] (policy.north) -- (data.south) node[midway, right]{Update};
\draw[line] (data.east) -- (model.west) node[midway, above]{};
\draw[line] (model) -- (function) node[midway, above]{$\hat{f}_{{\theta}^{*}}$};
\draw[line] (model) -- (sel) node[near end, right]{$\hat{f}_{{\theta}^{*}}$};
\draw[line] (function) -- (policy) node[near end, above]{$(\alpha^*,\beta^*)$} ;
\draw[->] (eval.south) to[bend left=15] (policy.south);
\draw[line] (eval) -- (best) node[midway, above]{Best};


\end{tikzpicture}

  }
  \caption{Framework for \underline{Opti}mization \underline{Proxy} based end-to-end \underline{N}eural \underline{A}rchitecture \underline{S}earch (\our).}
  \label{fig:framework}
\end{figure*}

\textbf{Predictor-guided sampling NAS.}  
Individually heuristic NAS methods formulate the search as the ``black box'' optimization problem. Typical methods utilize Evolutionary Algorithms (EA) \textcolor{gray}{\cite{sun2020automatically}}, Reinforcement Learning (RL) \textcolor{gray}{\cite{zoph2017neural, zoph2018learning}}, Bayesian Optimization 
 (BO) \textcolor{gray}{\cite{DBLP:conf/aaai/WhiteNS21/BANANA}}, and Local Search (LS) \textcolor{gray}{\cite{DBLP:conf/uai/WhiteNS21}} methods as the exploration strategies. 
Overall, the individually heuristic NAS methods explore architectures within a discrete domain and generally rely on individual evaluation, which raises concerns about efficiency. Target this, predictor-based NAS employ surrogate models as predictors along with encoding schemes \textcolor{gray}{\cite{DBLP:conf/nips/WhiteNNS20encoding,DBLP:conf/nips/YanWSH21,DBLP:conf/icml/WistubaP20,DBLP:conf/iclr/BakerGRN18}} to provide quick evaluation to the optimizer or sampling process. Notable works include \texttt{NPENAS} \textcolor{gray}{\cite{wei2022npenas}}, \texttt{BANANAS} \textcolor{gray}{\cite{DBLP:conf/aaai/WhiteNS21/BANANA}} and \texttt{BRP-NAS} \textcolor{gray}{\cite{dudziak2020brp}}. 
Zero-cost metrics also play a crucial role in assisting the optimizer as the predictor without requiring exhaustive training. The fundamental principle is to leverage saliency criteria, e.g., \texttt{Fisher}, \texttt{Grad Norm}, \texttt{Grasp}, \texttt{Jac. Cov.}, \texttt{SynFlow}, \texttt{SNIP} \textcolor{gray}{\cite{DBLP:conf/icml/MellorTSC21,DBLP:conf/iclr/AbdelfattahMDL21}}. 
However, according to the empirical study of \textcolor{gray}{\cite{colin2022adeeperlook}}, the nature of unreliable performance, harmful biases, and inconsistent performance across tasks of zero-cost metrics may have strong preferences that may bias the search, especially the data-agnostic proxies. \our\ focuses on improving search efficiency, it operates entirely in parallel with existing EA, RL, and BO approaches, yet is theoretically orthogonal and compatible with existing proxy models and zero-cost metrics.\\
\textbf{Weight-sharing NAS.} 
From another perspective, to address the evaluation-expensive issue, several studies \textcolor{gray}{\cite{pham2018efficient, Brock2017SMASH, bender2018understanding, guo2019single}} introduce the weight-sharing evaluation strategy by constructing a unified hypernetwork, from which sub-architectures shared the weights. 
Additionally, the \texttt{DARTS}-series methods \textcolor{gray}{\cite{liu2019darts, chen2019progressive, DBLP:conf/cvpr/LiQD0TG20, DBLP:conf/iclr/XieZLL19, DBLP:conf/cvpr/Xiao0Z0L22}} relax the architecture space, then optimize both the architecture parameters and supernetwork weights through gradient descent, significantly enhancing the efficiency of the search and evaluation. 
\our\ does not involve training acceleration strategies, in principle, they are orthogonal and compatible with our framework. We also draw on the concept of re-parameterization trick to reach the unbiased relaxation of the discrete architecture space in \texttt{SNAS} \textcolor{gray}{\cite{DBLP:conf/iclr/XieZLL19}}, however, not for the parameters learning of architectural weights in supernetwork, but for the relaxed architecture variable. \\
\textbf{Generative NAS.}
Inspired by the advancements of generative models, it offers a new perspective on reformulating the NAS problem, transforming the sampling-and-validation optimization process into a generative one, e.g., \texttt{GA-NAS} \textcolor{gray}{\cite{DBLP:conf/ijcai/RezaeiHNSMLLJ21}} \texttt{AG-Net} \textcolor{gray}{\citep{DBLP:journals/corr/abs-2203-08734}}, \texttt{DiffusionNAG} \textcolor{gray}{\cite{DBLP:journals/corr/abs-2305-16943}}. Especially, \texttt{DiNAS} \textcolor{gray}{\cite{DBLP:journals/corr/abs-2403-06020}} achieves the current state-of-the-art results across nearly almost NAS benchmarks.
Inspired by diffusion models \textcolor{gray}{\cite{DBLP:conf/cvpr/RombachBLEO22}} on visual generation tasks, \texttt{DiffusionNAG} \textcolor{gray}{\cite{DBLP:journals/corr/abs-2305-16943}} employs a diffusion model for their generation. The predictor-guided mechanism in \texttt{DiffusionNAG} enables the generation of task-specific optimal architectures by sampling from regions likely to exhibit desired properties. \texttt{DiNAS} \textcolor{gray}{\cite{DBLP:journals/corr/abs-2403-06020}} introduces a graph diffusion-based NAS method, utilizing discrete conditional diffusion processes to generate high-performing architectures. \\
\textbf{Optimization proxy.} As the proceeded of this work, we became aware of similar concepts of ``Smart Predict-Then-Optimize'' \textcolor{gray}{\cite{DBLP:journals/mansci/ElmachtoubG22}}, being applied to traditional optimization problems. Additionally, optimization proxies, particularly those based on deep neural networks, are designed to reduce the computational load of traditional solvers by directly predicting optimal or near-optimal solutions \textcolor{gray}{\cite{DBLP:journals/corr/abs-2311-13087}}. Common application scenarios include reinforcement learning in decision-making problems \textcolor{gray}{\cite{DBLP:journals/jair/YuanCH22}} and large-scale economic dispatch problems \textcolor{gray}{\cite{DBLP:journals/corr/abs-2304-11726}}. Independently conceived, we bring up this similar concept to formulate the NAS as a gradient-based ``End-to-end Predict-then-Optimize'' framework.
\subsection{Search framework}
\label{sec:frameworks}
For clarity and clear comparison with related works, we additionally present the schematic of the pipeline of \our\ framework in \textcolor{blue}{\pref{fig:framework}}.
\begin{figure*}[!t]
\centering
\subfloat[]{
\includegraphics[width=5.7cm]{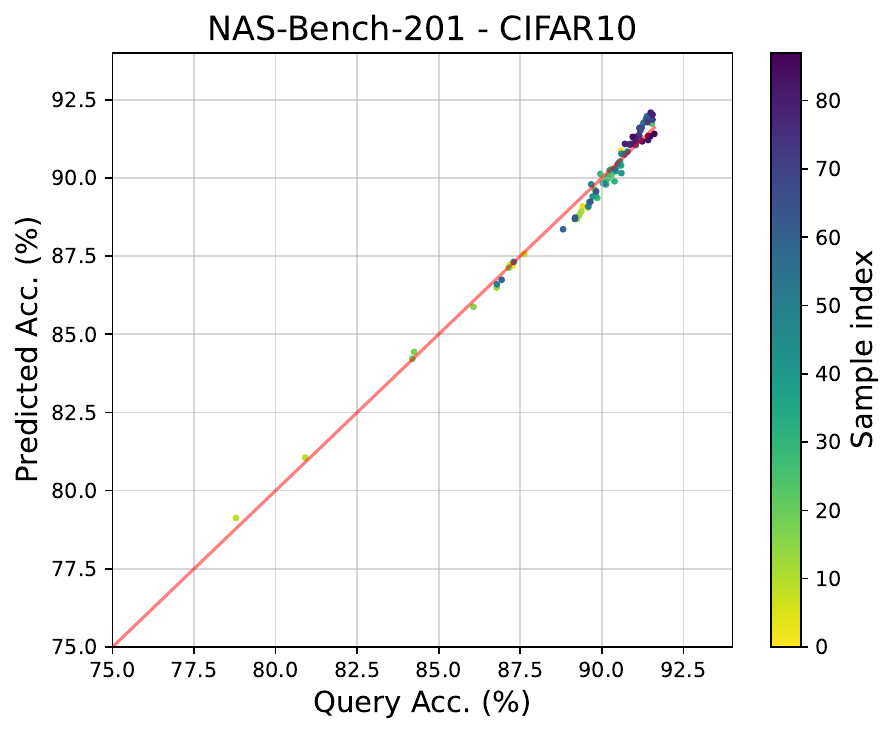}
}
\subfloat[]{
\includegraphics[width=5.7cm]{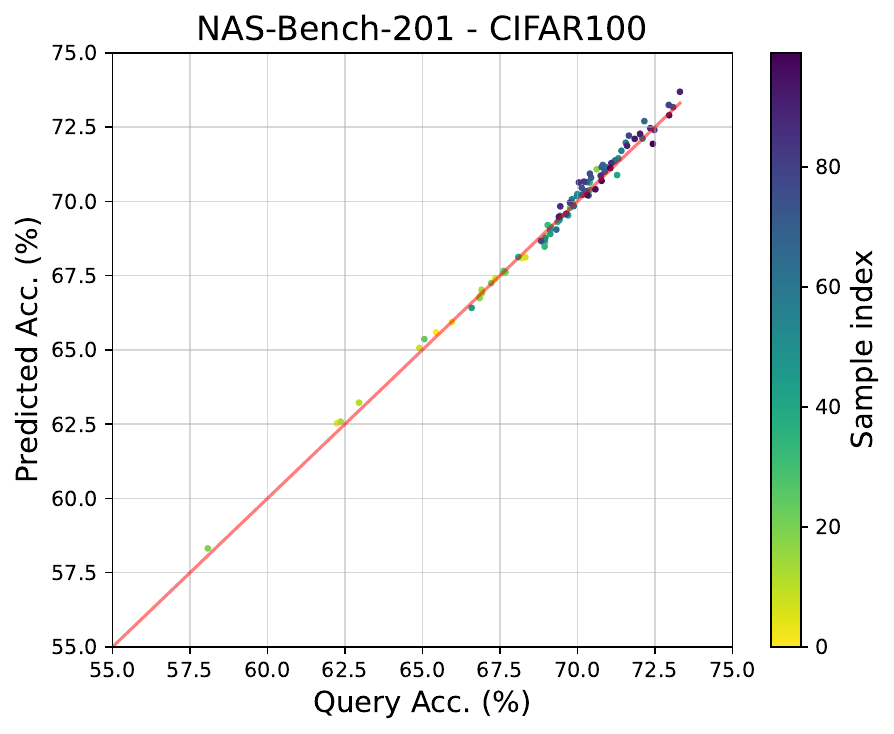}
}
\subfloat[]{
\includegraphics[width=5.7cm]{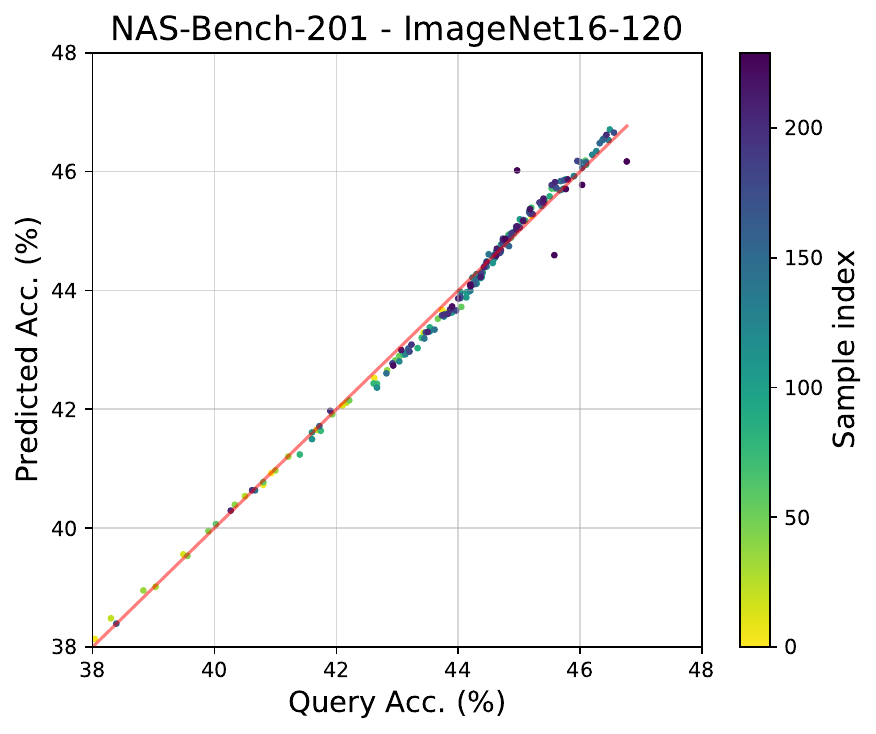}
}
\quad
\subfloat[]{
\includegraphics[width=5.7cm]{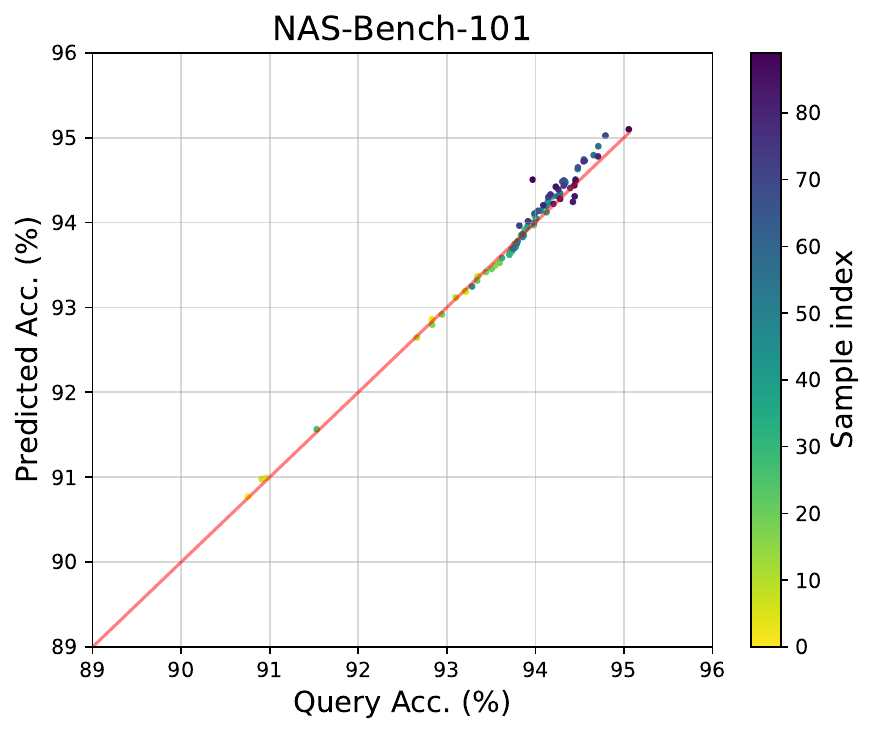}
}
\subfloat[]{
\includegraphics[width=5.7cm]{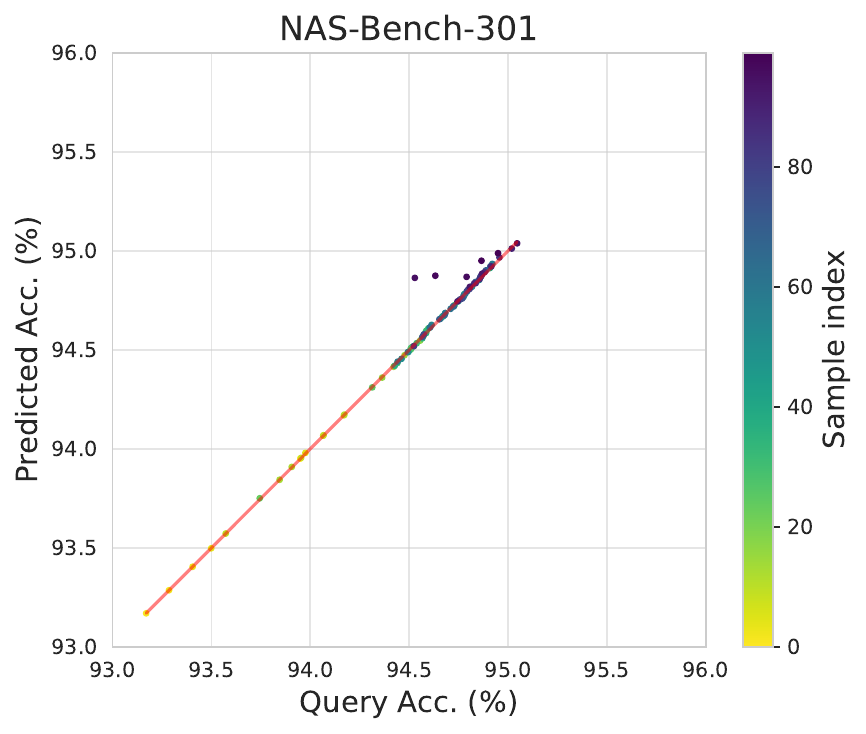}
}
\subfloat[]{
\includegraphics[width=5.7cm]{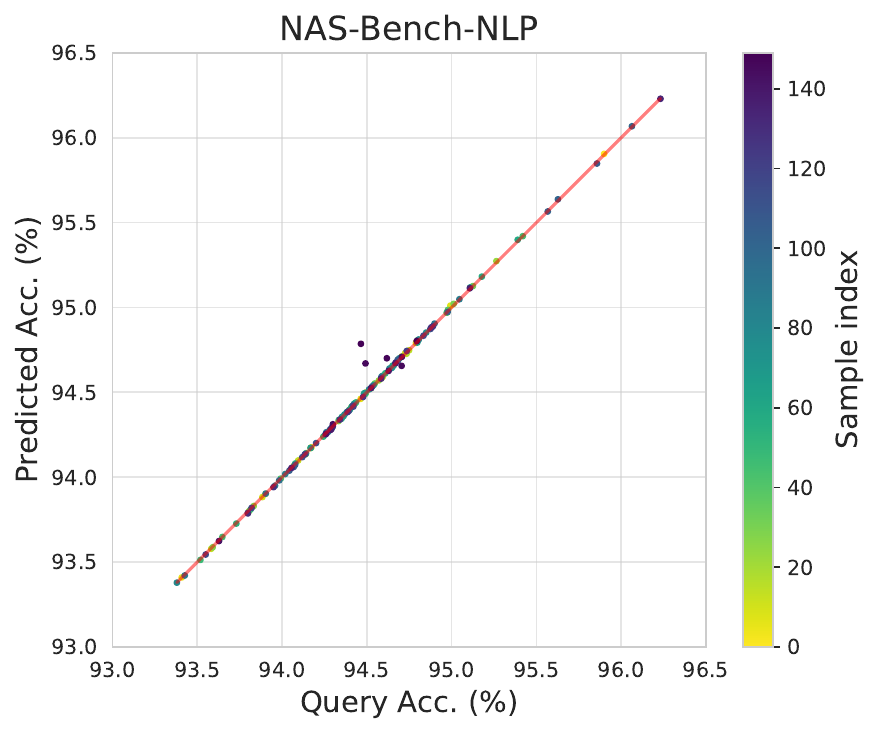}
}
\caption{The scatter of the evaluated architectures. We distinguish the sampling order of the architecture points by coloring them.}
\label{fig:pred_vs_true}
\end{figure*}
\subsection{Proxy models}
\label{sec:surrogatemodel}
We provide a detailed description of the GCN model, which we employ as the proxy model in our preliminary experiments. The initialization of each node in the graph begins with a $D_0$-dimensional representation, providing an initial feature vector \( V_0 \in \mathbb{R}^{N \times M} \), which is the one-hot vector to indicate the operation choice. The adjacency matrix \( A \in \mathbb{R}^{N \times N} \) represents the node connections, For each propagation layer $l$, a learnable weight matrix is as \( W_l \in \mathbb{R}^{D_l \times D_{l+1}} \), and layer-wise propagation as follows:
\begin{equation}
V_{l+1} = \text{ReLU}(A V_l W_l).
\end{equation}
The prior predictor-based NAS study \textcolor{gray}{\cite{wen2020neural}} employs the \texttt{average} operation over the bidirectional information propagation. One direction propagates information using \( A \) and another direction by \( A^T \), follows:
\begin{equation}
\begin{aligned}
{V}_{l+1}=1/2*\text{ReLU}({AV}_l{W}_l^+)+1/2*\text{ReLU}(A^{T} V_l W_l^-)
\end{aligned}
\end{equation}
Alternatively, we use the \texttt{concatenation} followed by a linear layer to preserve the information from both directions. Further, we add layer normalization to keep variables within a proper range and \texttt{GeLU} activation for better nonlinearity in the fully connected layer:
\begin{equation}
\begin{aligned}
V_{l+1} = \mathcal{LN} ( Linear(\text{concat}(&\text{GeLU}(A V_l W_l) + \\
&\text{GeLU}(A^T V_l W_l))) )
\end{aligned}
\end{equation}
We employ the weighted combination of the Mean Squared Error (MSE) loss and the pairwise ranking loss, as:
\begin{equation}
\begin{aligned}
\mathcal{L} = \mathcal{L}_{mse} + \lambda \mathcal{L}_{rank},
\end{aligned}
\end{equation}
where $\lambda$ is the ranking loss weight coefficient, and:
\begin{equation}
\begin{aligned}
&\mathcal{L}_{rank}(\{(\mathcal{A}_j, y_j)\}_{j=1, \dots, N}) = \\
&\sum_{j=1}^N \sum_{\{i, j | y_i > y_j\}} \max(0, m - (\hat{f}(\mathcal{A}_i) - \hat{f}(\mathcal{A}_j))).
\end{aligned}
\end{equation}
where $m$ is the margin, and $y_i$ is the ground truth accuracy of $\mathcal{A}_i$, this ranking loss emphasizes the relative ordering of predicted values rather than their absolute magnitudes. During the sampling process of our experiments, we monitor and record the distribution of model predictions and ground-truth values for the sampled and evaluated architectures. As shown in \textcolor{blue}{\pref{fig:pred_vs_true}}, it not only demonstrates the model's representation of the space landscape but also roughly illustrates the sampling process and convergence trend. Theoretically, our framework is compatible with most gradient-based models. We conduct the experiment with MLP, LSTM, and GAT on NAS-Bench-201, with results available in \textcolor{blue}{\pref{tab:gradien-based_nn}}. Since these are merely preliminary experiments, similar to the GCN model, we only employ the original encoding scheme, and stretch or serialize the $\mathcal{A}_t$ and $\mathcal{A}_o$ to adapt to different proxy models. Therefore, the experiments on these models are not comprehensive and in-depth.
\begin{table}[htb]
\centering
\caption{Preliminary experiments of different models on NAS-Bench-201 under 100 query times over 10 seeds ($0\sim9$).}
\resizebox{0.42\textwidth}{!}{
\label{tab:gradien-based_nn}
\begin{tabular}{@{}cccc@{}}
\toprule
Model& CIFAR-10 & CIFAR-100 & ImageNet16-120\\ \midrule
\textbf{Optimum*}       & $91.61$     & $73.49$     & $46.77$   \\ 
\midrule
GCN     & $91.61$     & $73.49$ & $46.64$\\
MLP     & $91.61$     & $73.49$ & $46.61$\\
LSTM    & $91.14$     & $70.51$ & $46.07$\\
GAT     & $91.44$     & $72.91$ & $45.81$\\
\bottomrule
\end{tabular}
}
\end{table}

\subsection{Proofs of relaxation unbias}
Before providing the proofs, we first reiterate the design intention of our search algorithm to ensure clarity. Here we elaborate our rationale from these aspects:
\begin{itemize}
    \item \textbf{Design Intention:} 
    The Bernoulli distribution in Eq. \eqref{eq:bernoulli} models the probability of connections between operations, enabling a continuous relaxation of the topological space. Although the assumption of edge independence simplifies the model, our experiments demonstrate that the proxy model effectively captures interactions or joint probability distributions between edges, enhancing the representation capability of our ``optimization proxy''.
    
    \item \textbf{Design Benefits:} 
    This approach simplifies the representation of binary decisions (i.e., whether connections exist or not). For each edge, only one parameter (i.e., the continuous probability $p$) needs to be optimized using gradient-based methods.
    
    \item \textbf{Co-design of Sampling and Proxy Sampling:} 

         For node features, the proxy sampling based on Eq. \eqref{eq:categorial} aligns with the discrete sampling of Eq. \eqref{eq:forward_ops} in the main paper part.
         
         For topological features, the proxy sampling based on Eq. \eqref{eq:bernoulli} aligns with the discrete sampling of Eq. \eqref{eq:forward_adj} in the main paper part.
    
    \item \textbf{Consistent Mathematical Form:} 
    The Bernoulli distribution is a special case of the categorical distribution for two outcomes, ensuring consistency in the mathematical form across related functions.
\end{itemize}

We present the proof of unbiased relaxation in operation node propagation in Eq. \eqref{eq:lambda_alpha} of the main paper part. For clarity, we omit the node index $i$, consider a specific feature node, and denote the operation index $j$ as the subscript, we denote Eq. \eqref{eq:lambda_alpha} in the main part as:
\label{sec:proof}
\begin{equation}
\label{eq:forward_ops_append}
\begin{aligned}
\hat{s}_{o}(\alpha)_{j} = \frac{\exp \left(\left( \alpha_{j}+g_{j}\right) / \tau\right)}{\sum_{m=1}^M \exp \left(\left( \alpha_{m}+g_{m}\right) / \tau\right)},
\end{aligned}
\end{equation}
where $0\leq j < M$, and \(\tau\) is the temperature. $\alpha_j$ is the unnormalized-logit of probability of the candidate operation choice. Further, $g$ follows Gumbel distribution, i.e., $g = -\log(-\log(u))$, where $u \sim \text{Uniform}(0, 1)$. Additionally, we have $\alpha = \log(\pi)$, where $\pi \in (0,\infty)$.
It has been proved in \textcolor{gray}{\cite{DBLP:conf/iclr/MaddisonMT17}} that:
\begin{equation}\label{eq:lambda_alpha_1}
\begin{aligned}
\mathbb{P}\left( \lim_{\tau \to 0} \hat{s}_{o}(\alpha)_{j} = 1 \right) & = \pi_j / {\sum_{m=1}^{M} \pi_m} = e^{\alpha_{j}}/{\sum_{m=1}^{M} e^{\alpha_{m}}} \\
& = \textit{softmax}(\alpha)_j,
\end{aligned}
\end{equation}
which is the categorical distribution in \textcolor{black}{Eq. \eqref{eq:categorial}} in the main body of this paper. 

For the unbiased relaxation in topological structure propagation, as Eq. \eqref{eq:lambda_beta} of the main paper part. We also omit the connection index $(h,k)$, i.e., considering a specific connection between two nodes, the Eq. \eqref{eq:lambda_beta} of the main paper part expressed as:
\begin{equation}\label{eq:gumbel_0}
\begin{aligned}
\hat{s}_{t}(\beta) & = \sigma ((\beta+g) / \tau) \\
& = \frac{1}{1 + \exp(-(\beta+g)/\tau)}.
\end{aligned}
\end{equation}
Further, we denote $\beta=log(\delta)$, and $\delta = \frac{\gamma}{1-\gamma}$.
Thus, $\beta \in (-\infty, \infty)$, $\delta \in (0, \infty)$, $\gamma \in (0,1)$, we have:
\begin{equation}\label{eq:gamma}
\begin{aligned}
\gamma=\sigma(\beta),
\end{aligned}
\end{equation}
and:
\begin{equation}\label{eq:gumbel_1}
\begin{aligned}
\hat{s}_{t}(\beta) = \frac{1}{1 + \exp(-((\log(\delta)+g)/\tau))}.
\end{aligned}
\end{equation}
Following the definition in \textcolor{gray}{\cite{DBLP:conf/iclr/MaddisonMT17}}, $\hat{s}(\beta) \in (0, 1)$ follows a Binary Concrete distribution, i.e., $\hat{s}(\beta) \sim \text{BinConcrete}(\delta, \tau)$ with location $\delta$ and temperature $\tau$.
As demonstrated in \textcolor{gray}{\cite{DBLP:conf/iclr/MaddisonMT17}}, we have:
\begin{equation}\label{eq:gumbel}
\begin{aligned}
\mathbb{P}( \lim_{\tau \to 0} \hat{s}_{t}(\beta) & = 1 ) = \delta / (1+\delta) \\
& = \frac{\gamma/(1-\gamma)}{1+\gamma/(1-\gamma)}
& = \gamma \\
& = \sigma(\beta),
\end{aligned}
\end{equation}
which is the Bernoulli distribution variables in \textcolor{black}{Eq. \eqref{eq:bernoulli}} of the main body of this paper.

\subsection{Notations summary}
\label{sec:notation}
We provide a comprehensive summary of the notations used throughout this paper in\textcolor{blue}{~\pref{tab:notation}}, facilitating a clear understanding of the mathematical formulations and concepts presented.
\begin{table}[h!]
\centering
\caption{Notation Table}
\label{tab:notation}
\resizebox{0.45\textwidth}{!}{%
\begin{tabular}{|c|m{6cm}|}
\hline
\textbf{Symbol} & \textbf{Description} \\ 
\hline
$\mathbb{S}$ & The discrete NAS search space \\ \hline
$\mathcal{X}$ & The continuously relaxed NAS search space spanning by $\alpha$ and $\beta$, proxy space \\ \hline
$\mathcal{A}$ & Architecture in $\mathbb{S}$ \\ \hline
$\alpha$ & The continuous operation feature variable of architecture (the unnormalized-logit value of the probability of the candidate operation) \\ \hline
$\beta$ & The continuous topological structure of architecture (the Log-odds value of the connection probability between operation nodes) \\ \hline
$\gamma$ & Connection probability between operation feature nodes \\ \hline
$\delta$ & The odds, how likely the connection exists \\ \hline
$g$ & The i.i.d samples drawn from Gumbel distribution \\ \hline
$\tau$ & Temperature parameter in Gumbel softmax distribution \\ \hline
${f}$ & Landscape function \\ \hline
$\hat{f}$ & Proxy model function \\ \hline
$\theta$ & Weights or parameters of the proxy model \\ \hline
$\theta^*$ & Optimal weights or parameters of the proxy model \\ \hline
$\mathcal{A}^*$ & Optimal architecture \\ \hline
$\mathcal{A}_o, \mathcal{A}_t$ &  The topological structure and features variables (candidate operation choices) \\  \hline
$\mathcal{L}$ & Loss function\\ \hline
$\mathcal{D}$ & Sampled dataset $\{(\mathcal{A}_i, f(\mathcal{A}_i))\}_{i=1}^n$ \\ \hline
$s$ & Sampling process from the continuous probability distribution to discrete architecture \\ \hline
$\hat{s}$ & Proxy sampling function \\ \hline
$\hat{s}_o, \hat{s}_t$ &  Soft embeddings of the topological structure and features variables \\   \hline
$i,j$ & Feature nodes index and candidate operation index \\ \hline
$h,k$ & Current and another feature nodes index  \\ \hline
$\psi(\cdot)$ & Objective function \\ \hline
$\sigma(\cdot)$ & Sigmoid function \\ \hline
$\text{Cat}(\cdot)$ & Categorical distribution \\ \hline
$\text{Bernoulli}(\cdot)$ & Bernoulli distribution \\ \hline
$N$ & The No. of feature nodes \\ \hline
$M$ & The candidate operation number \\ \hline
$C$ & The maximum query time cost \\ \hline
$Q$ & The sampled candidate architectures number of each search step \\ \hline
$B$ & The evaluation (quer) architecture number of each search step  \\ \hline
$\lambda$ & The ranking loss weight coefficient  \\ \hline
\end{tabular}
}
\end{table}


\subsection{Experiments}
\label{sec:exp_details}


\begin{figure*}[!htb]
\centering
\subfloat[]{
\includegraphics[width=5.7cm]{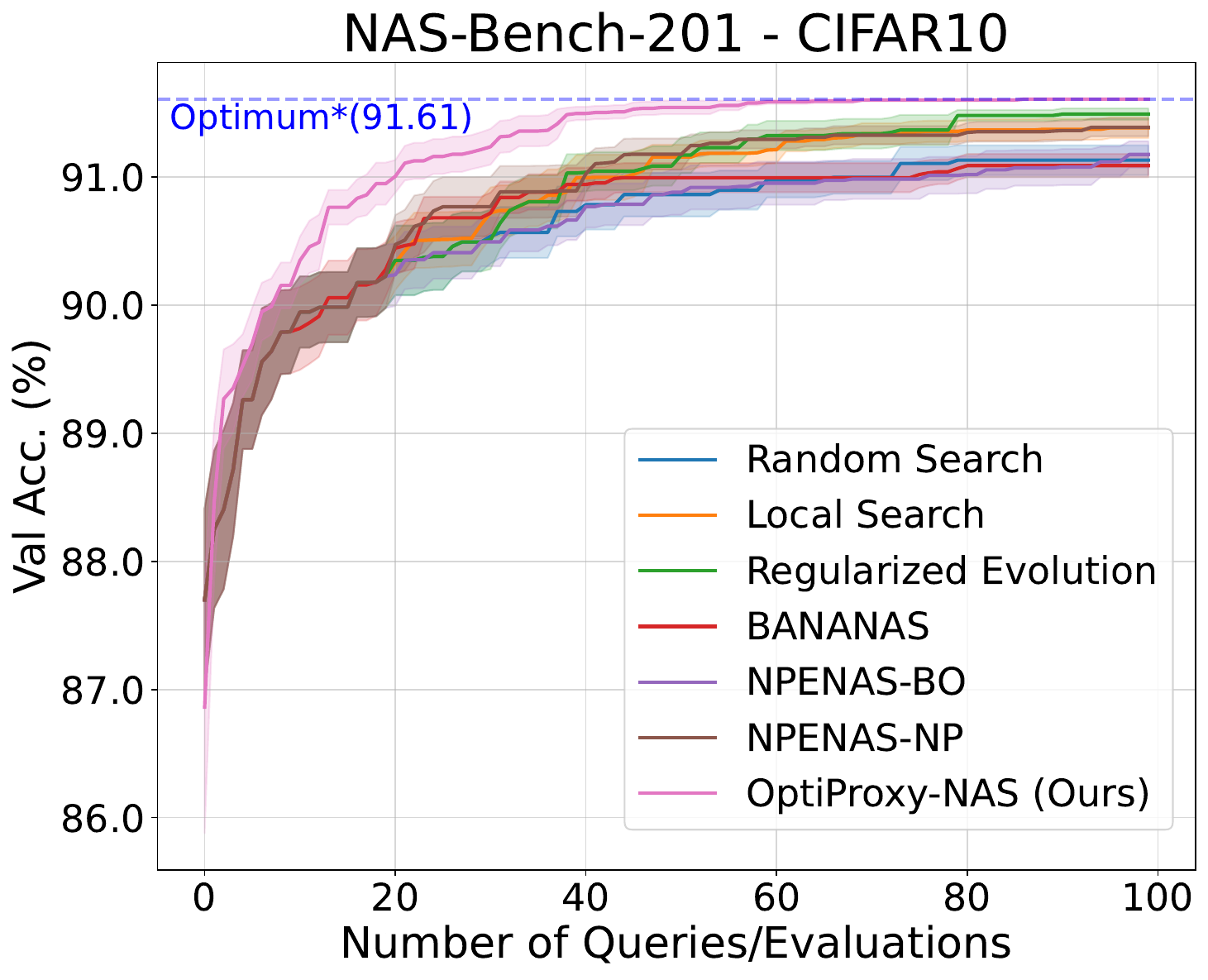}
}
\subfloat[]{
\includegraphics[width=5.7cm]{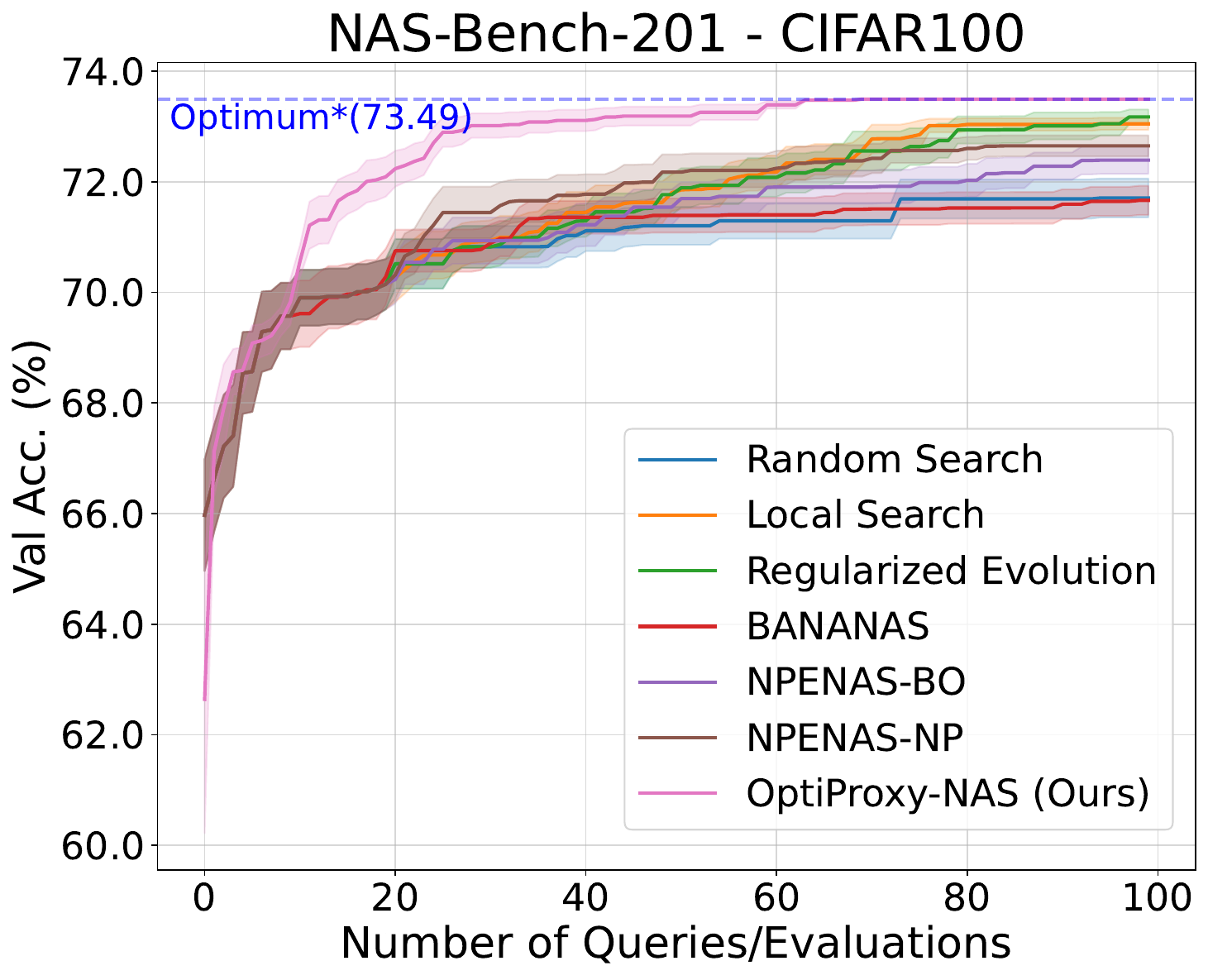}
}
\subfloat[]{
\includegraphics[width=5.7cm]{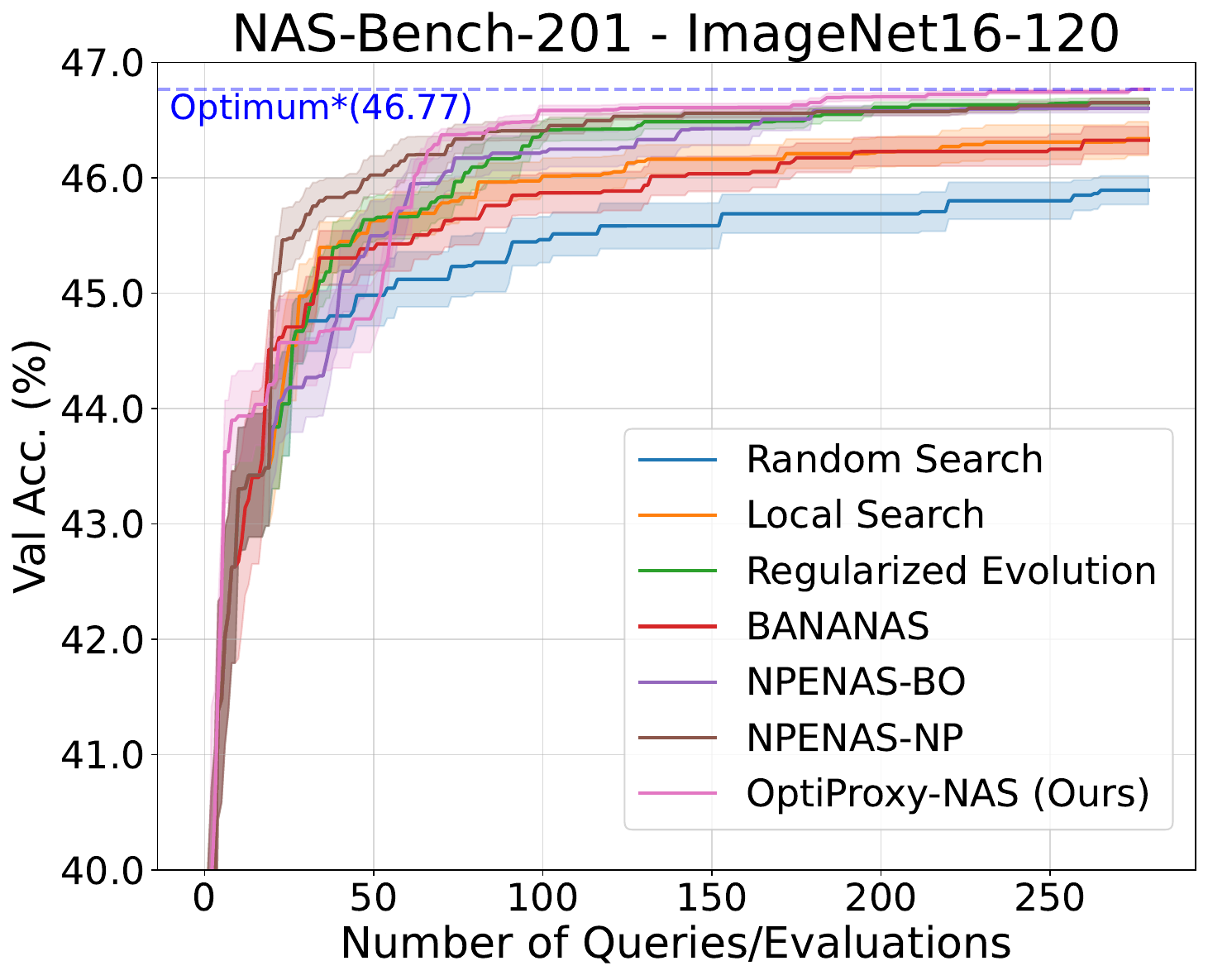}
}
\quad
\subfloat[]{
\includegraphics[width=5.7cm]{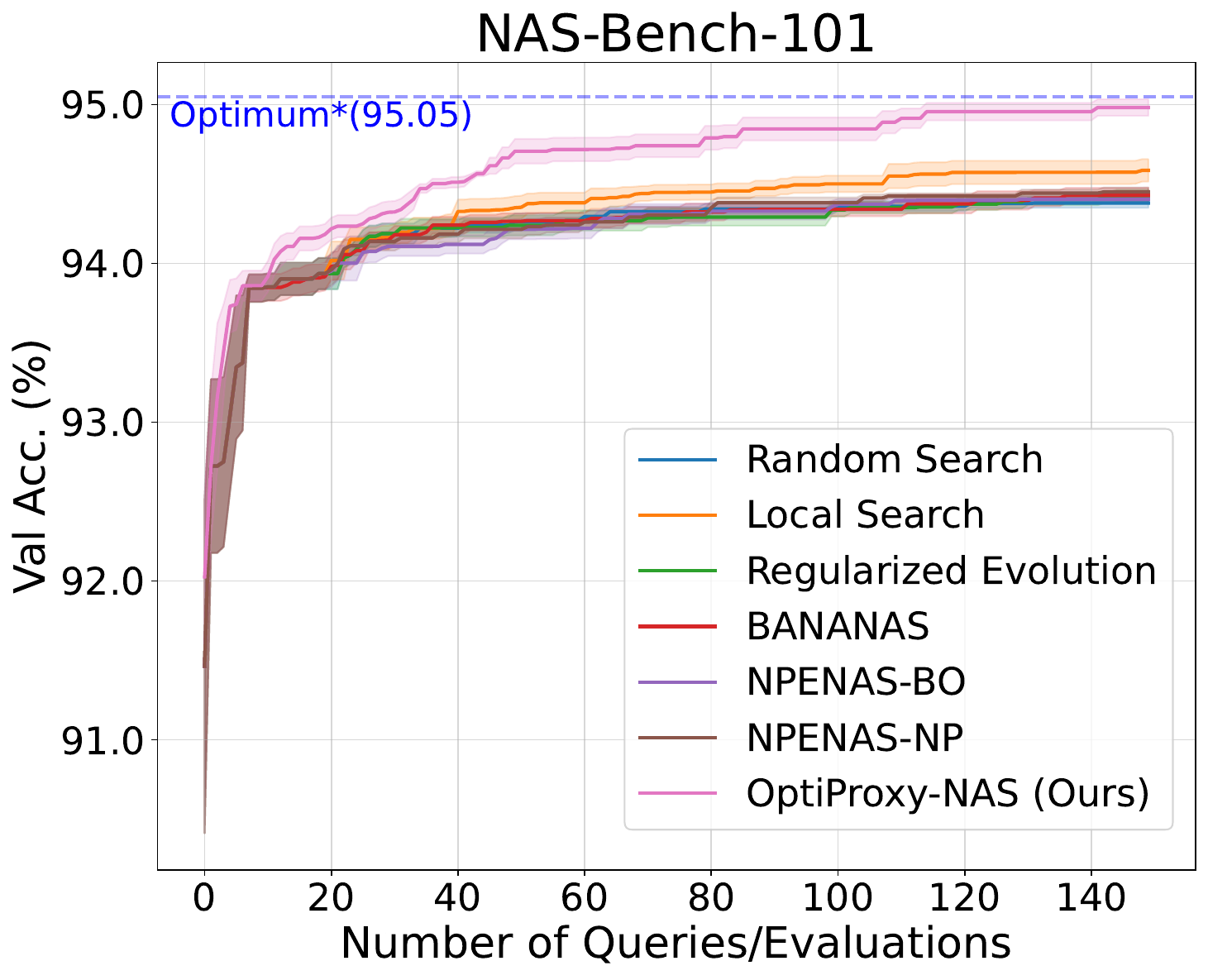}
}
\subfloat[]{
\includegraphics[width=5.7cm]{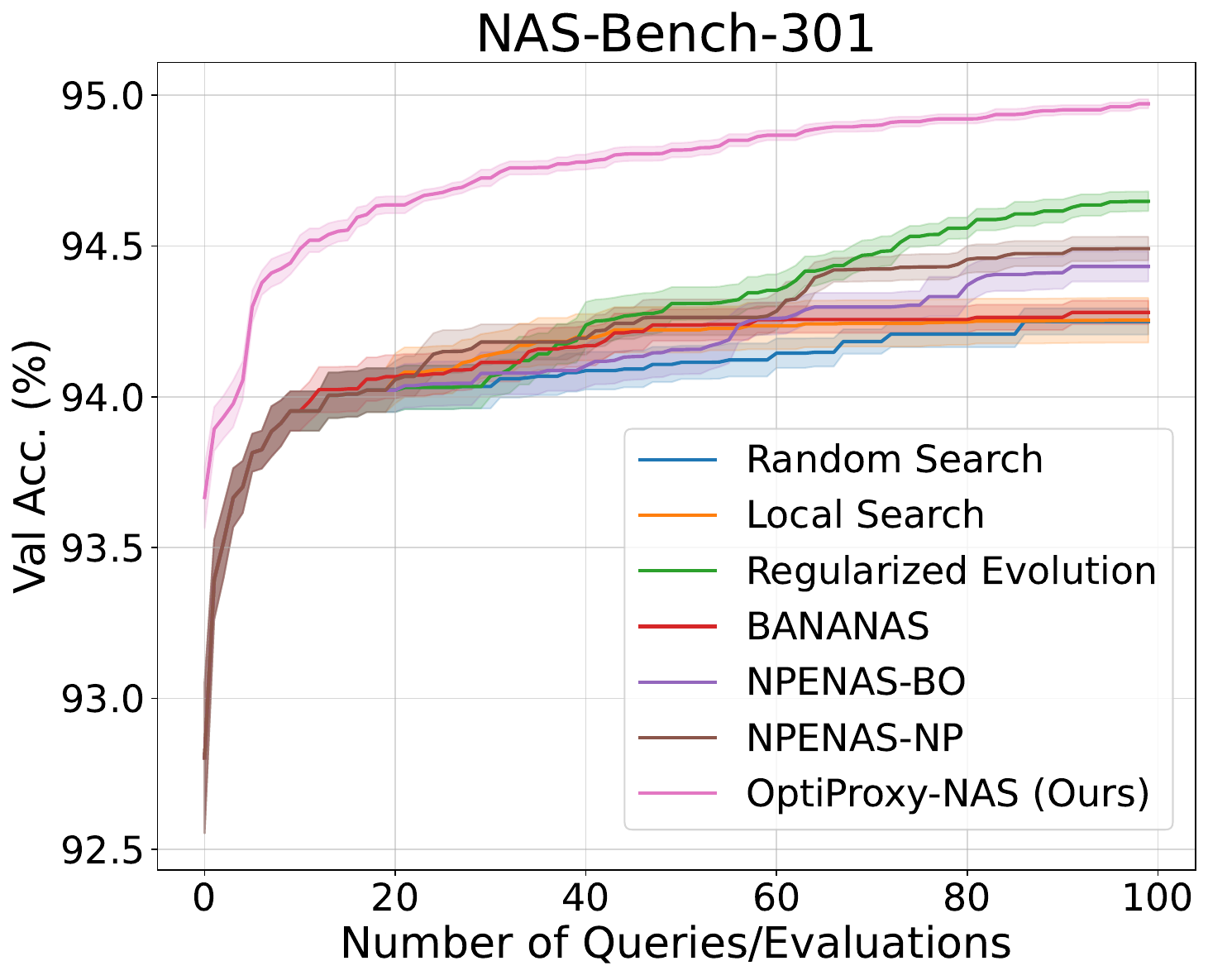}
}
\subfloat[]{
\includegraphics[width=5.7cm]{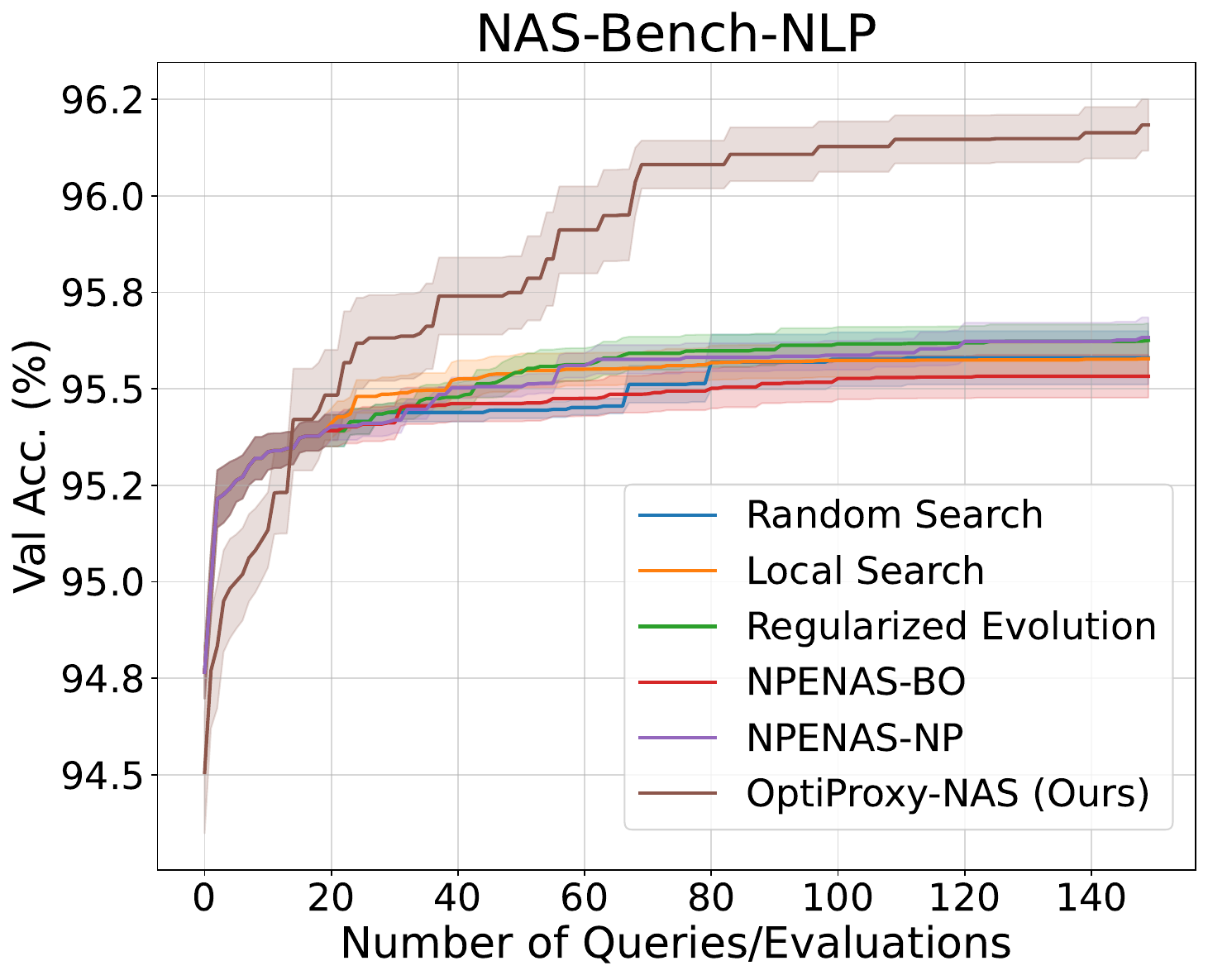}
}
\caption{The trajectories in each subplot illustrate the statistics of the best-so-far values versus the query number across $10$ random seeds. The solid lines represent the mean values, and the shaded areas indicate the standard error.}
\label{fig:query_efficient}
\end{figure*}
\subsubsection{Experiment setup}
\label{sec:setup}
Our detailed experimental setup is as follows:
\begin{itemize}
    \item Benchmarks: Following the standard practice in NAS algorithm comparisons, we choose the well-known benchmark datasets, including NAS-Bench-101, NAS-Bench-201, NAS-Bench-301, NAS-Bench-NLP, and HW-NAS-Bench, which spans {$7$} tasks of {$4$} search spaces across {$3$} different domains including computer vision, natural language processing, and resource-constrained NAS. For the types and structure of proxy models (neural networks), we employ the \texttt{XGBoost} as the surrogate model, also consistent with other comparison methods, e.g., \texttt{AG-Net} and \texttt{DiNAS}. Related versions are also maintained consistently: $\blacktriangleright$ NAS-Bench-201: V$1.0$, $\blacktriangleright$ NAS-Bench-301: V$0.9$; $\blacktriangleright$ HW-NAS-BENCH: V$1.0$. In addition, surrogate model choice of surrogate benchmarks: $\blacktriangleright$ NAS-Bench-301: \texttt{XGBoost}; $\blacktriangleright$ NAS-Bench-NLP: \texttt{svd\_lgb}.
    \item Comparison methods: We choose the SOTA methods in terms of sample efficiency under the mainstream search strategies, details in \textcolor{blue}{\pref{tab:nas_algorithms}}.
    \item Comparison metrics: Following convention, we focus on the accuracy of the best architecture obtained (e.g., search based on validation accuracy) and queries cost, which directly determine the expenses of the entire search pipeline. We conduct experiments with $10$ different seeds ($0\sim9$) and report the mean and standard deviation values (comparison methods typically report only the mean value).
    \item Dataset setting: The settings for the benchmark dataset include the version number, the search metric (\texttt{val}), whether to use the mean of the multi-seed training results or a single-seed result as the metric, and other settings. We ensure alignment with the latest SOTA comparison methods, \texttt{AG-Net} and \texttt{DiNAS}.
    \item Reproducibility: Following the comparative methodology, all experiments run with the random seeds from $0$ to $9$. We present the detailed reproducibility settings in the Appendix \textcolor{blue}{~\pref{sec:repreoduce}}.
\end{itemize}

\begin{table}[h]
\centering
\caption{Comparison of NAS Algorithm Categories}
\resizebox{0.45\textwidth}{!}{%
\begin{tabular}{|l|m{6cm}|}
\hline
\textbf{Category}           & \textbf{Comparison Methods}      \\ \hline
Random Search (RS)          &  \textcolor{gray}{\cite{DBLP:conf/uai/LiT19}}      \\ \hline
Evolutionary Algorithms (EA)&  Regularized Evolution \textcolor{gray}{\cite{DBLP:conf/uai/LiT19}}                  \\ \hline
Reinforcement Learning (RL) &  Arch2vec+RL \textcolor{gray}{\cite{DBLP:conf/nips/YanZAZ020}}                              \\ \hline
Local Search (LS)           & \textcolor{gray}{\cite{DBLP:conf/uai/WhiteNS21}}    \\ \hline
Predictor-based NAS          & WeakNAS \textcolor{gray}{\cite{DBLP:conf/nips/WuDCCLYWLCY21}} \\ \hline
Supernet-based NAS          & SGNAS \textcolor{gray}{\cite{DBLP:conf/cvpr/HuangC21}}, $\beta$-DARTS \textcolor{gray}{\cite{DBLP:conf/cvpr/YeL00FO22}}  \\ \hline
Generative NAS              & NAO \textcolor{gray}{\cite{DBLP:conf/nips/LuoTQCL18}}, GA-NAS \textcolor{gray}{\cite{DBLP:conf/ijcai/RezaeiHNSMLLJ21}}, DiffusionNAG \textcolor{gray}{\cite{DBLP:journals/corr/abs-2305-16943}}, DiNAS \textcolor{gray}{\cite{DBLP:journals/corr/abs-2403-06020}}, AG-Net \textcolor{gray}{\cite{DBLP:journals/corr/abs-2203-08734}} \\ \hline
Bayesian Optimization (BO)  & BANANAS$^{\dagger}$ \textcolor{gray}{\cite{DBLP:conf/aaai/WhiteNS21/BANANA}}, Bayesian Opt. \textcolor{gray}{\cite{DBLP:conf/icml/SnoekRSKSSPPA15}}, Arch2vec+BO \textcolor{gray}{\cite{DBLP:conf/nips/YanZAZ020}}   \\ \hline
\end{tabular}
}
\label{tab:nas_algorithms}
\end{table}

\subsubsection{Query efficiency experiment}
\label{sec:query}
We present the query efficiency experiment in \textcolor{blue}{\pref{fig:query_efficient}}. We choose the baseline methods that are incorporated in repo: \textcolor{blue}{https://github.com/automl/NASLib}, under the compatible search space, encoding scheme, and prediction models. Specifically, for tabular benchmarks (NAS-Bench-101 and NAS-Bench-201), our method demonstrates superior query efficiency.
For the surrogate benchmarks (NAS-Bench-301 and NAS-Bench-NLP), we consistently outperform the baseline methods along with the increase in query numbers. Specifically, we ignore the running of \texttt{BANANAS} on NAS-Bench-NLP due to its incompatible path encoding scheme.

\subsubsection{Visualization}
\label{sec:visual}
\begin{figure}[!ht]
\centering
\subfloat[]{
\includegraphics[width=6cm]{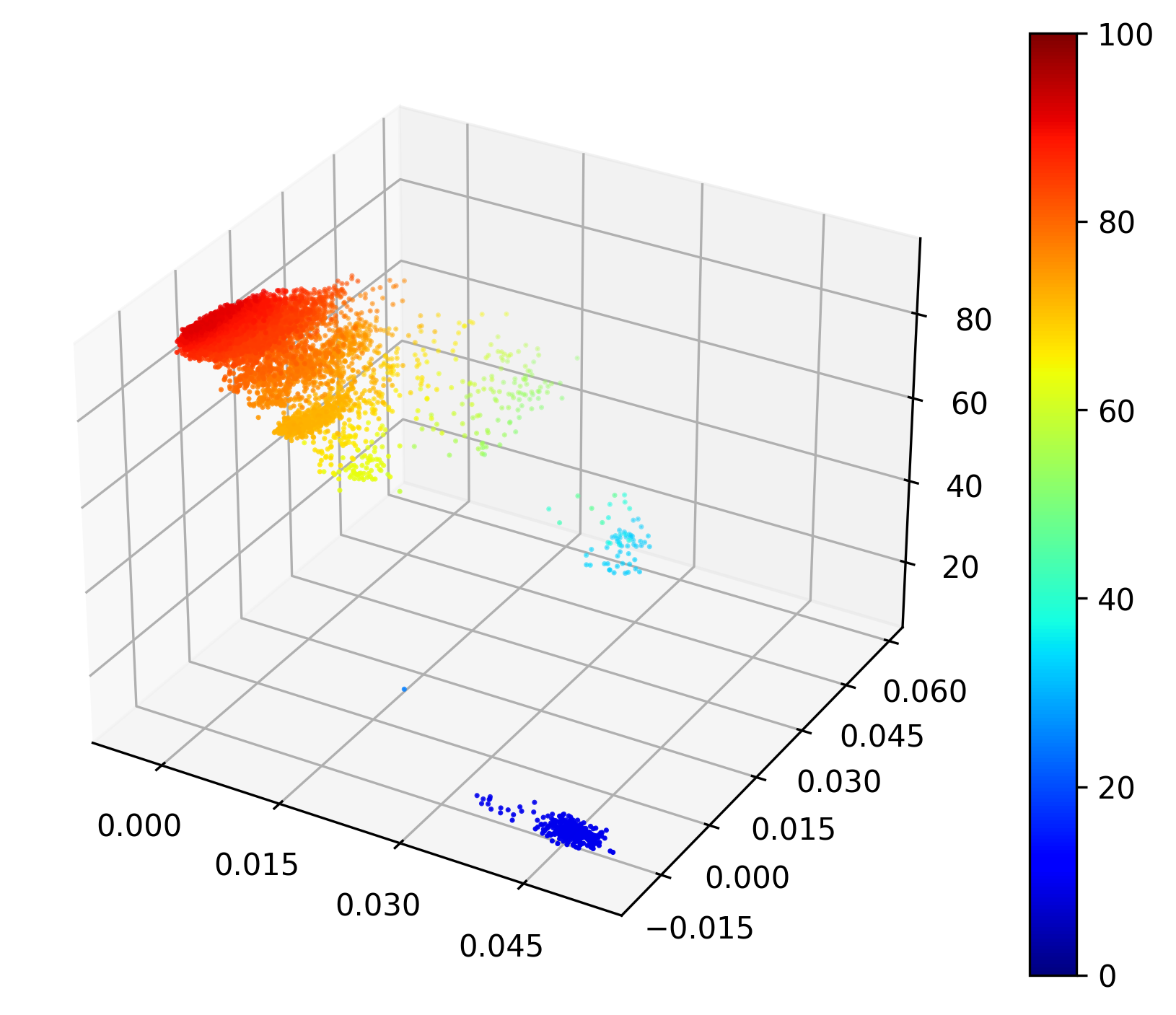}
\label{fig:overview}
}
\quad
\subfloat[]{
\includegraphics[width=6cm]{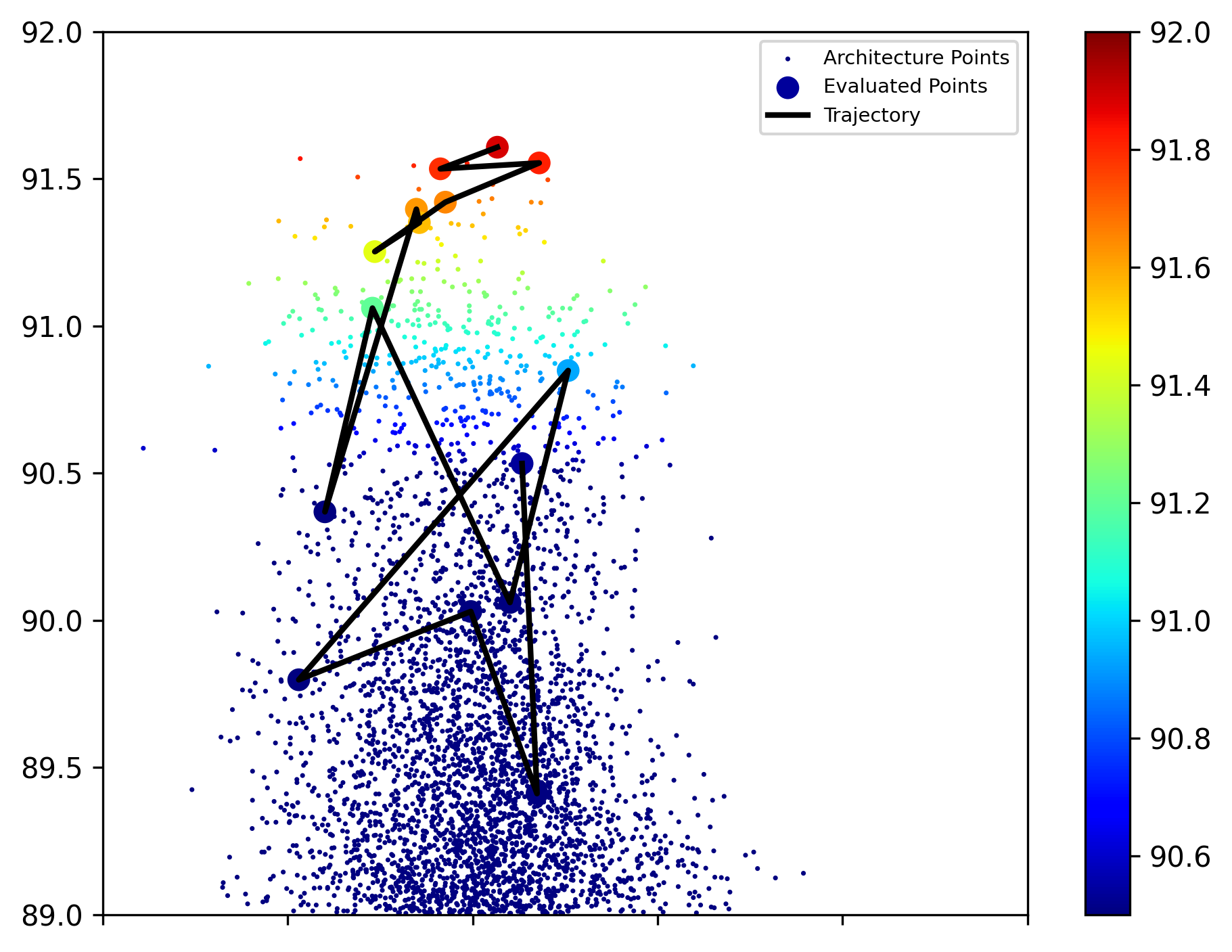}
\label{fig:zoom_in}
}
\quad
\subfloat[]{
\includegraphics[width=6cm]{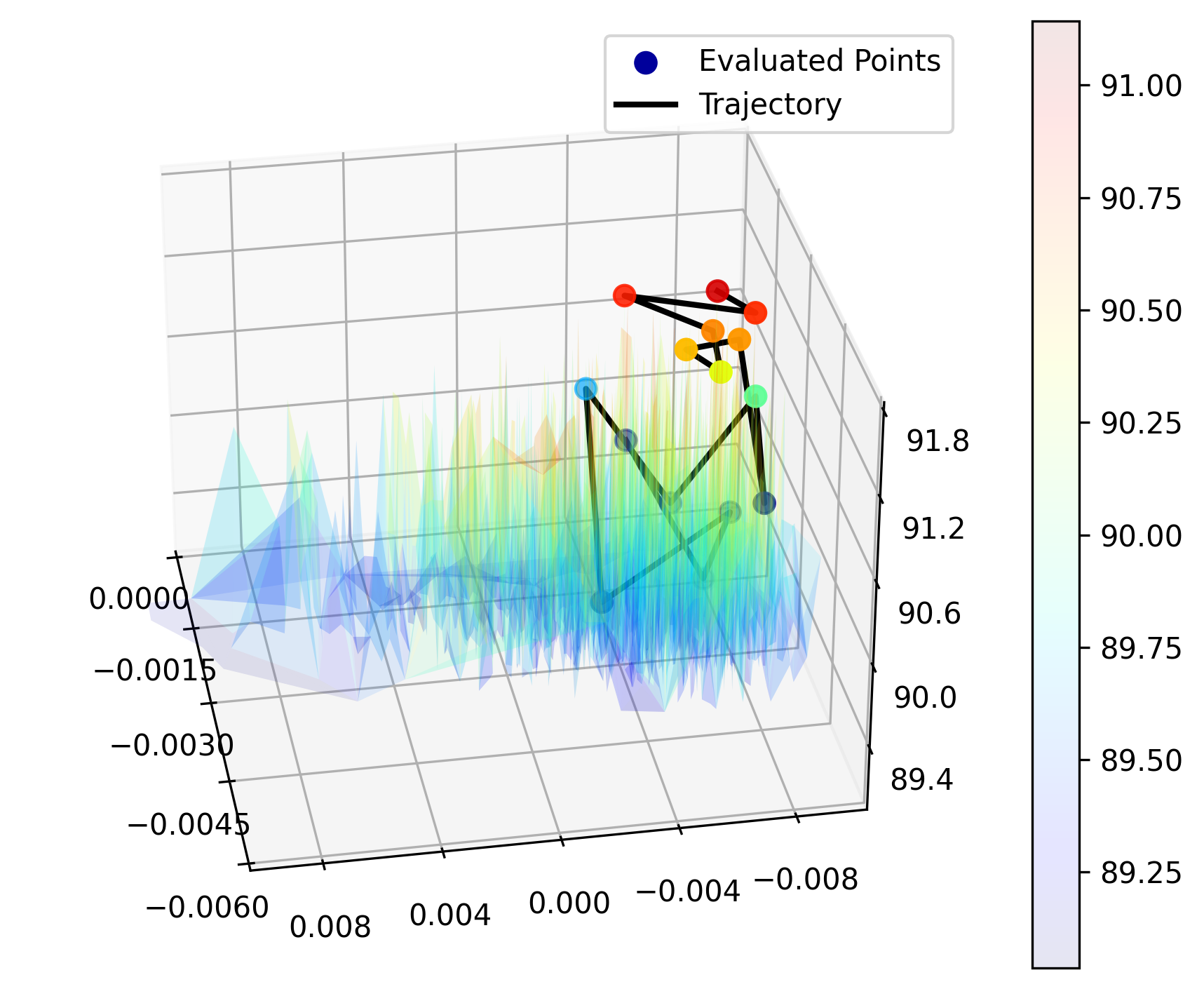}
\label{fig:surface}
}
\caption{Visualization of the search space architecture distribution and search trajectory on NAS-Bench-201 (CIFAR-10). (a) Overview of the search space using PCA. (b) The primary exploration region of our method, viewed from the $y$-direction, with a specific area zoomed in for detail. For clarity, only the best architecture point out of every five evaluated architectures is visualized. This illustration search takes $70$ queries to reach the global optimum. Based on our observations, every $10$ evaluations yield a new best architecture. (c) The search trajectory within the spiky search space landscape.}
\label{fig:visual}
\end{figure}
We visualize the search space of NAS-Bench-201, CIFAR-10 to demonstrate the properties of the search space and the effectiveness of the proposed method using the search trajectory, as \textcolor{blue}{~\pref{fig:visual}}. We perform PCA on these embeddings to reduce the dimensionality from $5$ to $2$ and add the validation accuracy as the third dimension to further differentiate the architectures based on their performance, as shown in \textcolor{blue}{~\pref{fig:overview}}.
Typically, our method demonstrated significant efficiency during the first few evaluations, the search could be led to the advantageous region quickly. So we zoom in on this region (the region of red points) for a closer look. To better differentiate the performance of the architectures in this region we use a more focused color-bar in this graph and view the space in $-y$ direction and the three-dimensional space landscape visualization, as \textcolor{blue}{\pref{fig:zoom_in}} and \textcolor{blue}{\pref{fig:surface}}, respectively. 

\subsection{Search Spaces Parameterization and Encoding schemes}
\label{sec:space_encoding}
\subsubsection{NAS-Bench-201.}
\begin{figure}[htb]
  \centering
    \subfloat[][\label{fig:201-graph}]{
      \includegraphics[width=0.32\columnwidth]{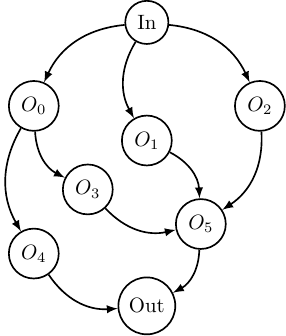}
    }
    \subfloat[][\label{fig:201-adj}]{
      \includegraphics[width=0.4\columnwidth]{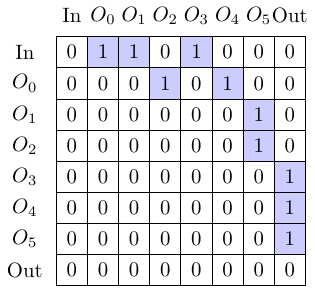}
    }
    \quad
    \subfloat[][\label{fig:201-alpha}]{
      \includegraphics[width=0.4\columnwidth]{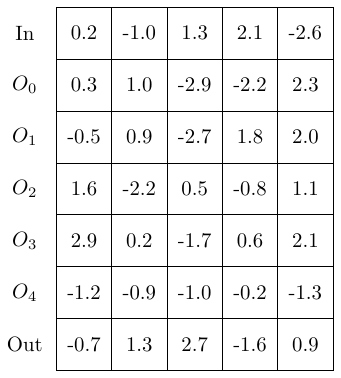}
    }
    \subfloat[][\label{fig:201-alpha-sample}]{
      \includegraphics[width=0.38\columnwidth]{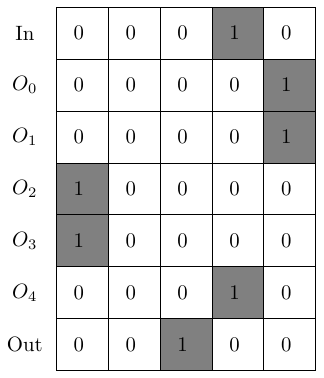}
    }
  \caption{(a) Illustration of search space of NAS-Bench-201. (b) The fixed topological structure (adjacency matrix) of the architecture cell. (c) The relaxed architecture operation feature variables $\alpha$. (d) The hard embedding of the topological structure that is sampled architecture based on $\alpha$. }
  \label{fig:201}
\end{figure}
\begin{figure}[!ht]
  \centering
  \subfloat[][\label{fig:201-constrain-0}]{
    \includegraphics[width=0.8\columnwidth]{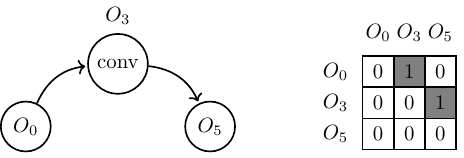}
  }
    \quad
  \subfloat[][\label{fig:201-constrain-1}]{
    \includegraphics[width=0.8\columnwidth]{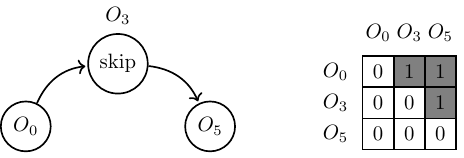}
  }
  \quad
  \subfloat[][\label{fig:201-constrain-2}]{
    \includegraphics[width=0.8\columnwidth]{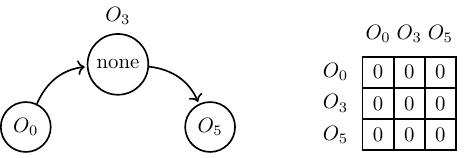}
  }
\caption{From the perspective of graph structure, illustration of the coupling relationship between operation choice and adjacency matrix. (a) The original connectivity pattern when the operation choice is ``conv\_3x3''. (b) When the ``skip\_connect'' is chosen, the edge originating from $O_3$ and connecting to $O_3$ will both be removed, and a new edge is additionally added to bypass the node $O_3$, resulting in the change of the adjacency matrix. (c) When the ``none'' is chosen, the edge originating from $O_3$ and connecting to $O_3$ will both be removed, resulting in the zero out of the adjacency matrix element.}
\label{fig:201-constraint}
\end{figure}
We present the illustration of the NAS-Bench-201 space and encoding scheme in \textcolor{blue}{\pref{fig:201}}.
The architecture in NAS-Bench-201 space has a fixed topological structure, as shown in \textcolor{blue}{\pref{fig:201-graph}}. The NAS targets the operation choice from the candidate operations set, with the entire space comprising $6^5$=$15,625$ architectures. The embedding scheme of NAS-Bench-201 can be summarized: $\blacktriangleright$ The $\alpha$ is as \textcolor{blue}{~\pref{fig:201-alpha}}, within the range of $\mathbb{R}$. $\blacktriangleright$ For hard embedding after the sampling $s$, adjacency matrix and operation feature are as \textcolor{blue}{\pref{fig:201-adj}} and \textcolor{blue}{\pref{fig:201-alpha-sample}}, respectively.
The candidate operation set in NAS-Bench-201 space is as follows:
\begin{minted}{Python}
opname_to_index = {
    'none': 0, 
    'skip_connect': 1, 
    'conv_1x1': 2, 
    'conv_3x3': 3, 
    'avg_pool_3x3': 4
}
\end{minted}
We further analyze the impact of operations on the topological structure. Given the original \textcolor{blue}{\pref{fig:201-constrain-0}}, we further discuss the vibrations over the operation selection change: 
$\blacktriangleright$ When ``skip\_connect'' is chosen, it changes the topological structure, represented as an adjacency matrix, as illustrated in \textcolor{blue}{\pref{fig:201-constrain-1}}.
$\blacktriangleright$ When ``none'' is chosen, the adjacency matrix is also affected, as illustrated in \textcolor{blue}{\pref{fig:201-constrain-2}}.
 We consider that these coupling factors between the operation matrix and the adjacency matrix can be neglected, that is, the adjacency matrix of this architectural space always exists in a fixed form as\textcolor{blue}{~\pref{fig:201-adj}}, which is also the commonly employed encoding method for this space in the NAS community.

\subsubsection{NAS-Bench-101.}
\label{sec:appendix_101}
\begin{figure}[!ht]
  \centering
  \subfloat[][\label{fig:101-graph}]{
    \includegraphics[width=0.32\columnwidth]{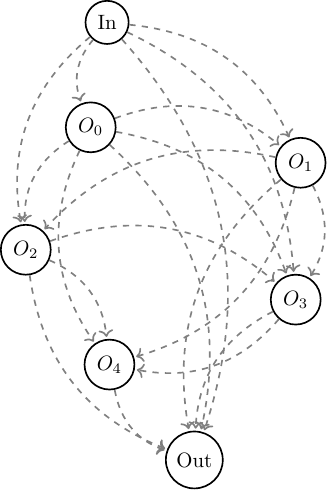}
  }
  \subfloat[][\label{fig:101-adj-c}]{
    \includegraphics[width=0.45\columnwidth]{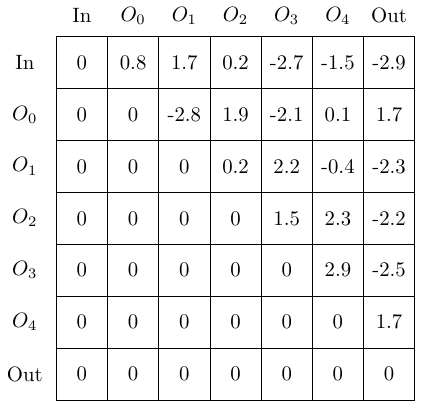}
  }
\quad
  \subfloat[][\label{fig:101-adj-d}]{
    \includegraphics[width=0.45\columnwidth]{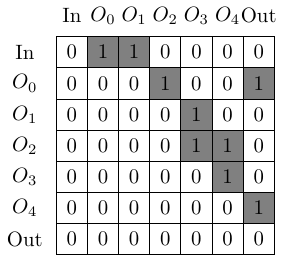}
  }
  \subfloat[][\label{fig:101-graph-1}]{
    \includegraphics[width=0.32\columnwidth]{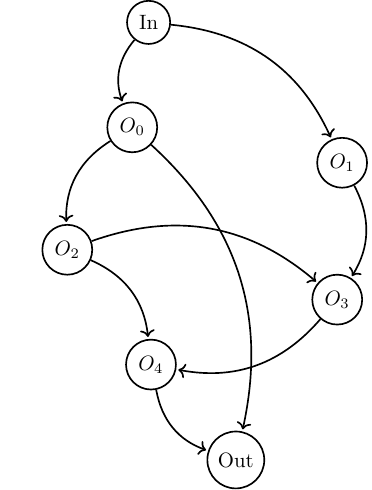}
  }

  \caption{Illustration of the search space of NAS-Bench-101. (a) The architecture cell of NAS-Bench-101. Both topological structure (edges represented by dashed line) and operation features (nodes) are to be searched. (b) $\beta$ in continuous space. (c) The encoding scheme (hard embedding) of graph topology structure that is sampled based on $\beta$. (d) The graph of sampled candidate architecture.}
  \label{fig:101}
\end{figure}
We present the illustration of the NAS-Bench-101 space and encoding scheme in\textcolor{blue}{~\pref{fig:101}}, both topological structure and operation features are to be searched. The embedding scheme of NAS-Bench-101 can be summarized: $\blacktriangleright$ The $\alpha$ is as the type of \textcolor{blue}{~\pref{fig:201-alpha}} of NAS-Bench-201, and the form of $\beta$ is as \textcolor{blue}{~\pref{fig:101-adj-c}}, each element is within the range of $\mathbb{R}$. $\blacktriangleright$ For hard embedding after the sampling $s$, the adjacency matrix is as\textcolor{blue}{~\pref{fig:101-adj-d}}.
The candidate operation set of NAS-Bench-101 space is as follows:
\begin{minted}{Python}
opname_to_index = {
    'conv3x3-bn-relu': 0,
    'conv1x1-bn-relu': 1,
    'maxpool3x3': 2
}
\end{minted}
Critically, the NAS-Bench-101 space imposes the following validity constraints on the architecture: 
$\blacktriangleright$ The edge number $\leq 9$.
$\blacktriangleright$ At least one emanates from an \texttt{In} node.
$\blacktriangleright$ At least one node outputs to the \texttt{Out} node.
Additionally, due to the collaborative changes and coupling relationship of the topological structure and node features in network architecture, isomorphic graphs exist in the space \textcolor{gray}{\cite{ying2019bench}}: 
$\blacktriangleright$ The two architectures have different adjacency matrices or node features, but are computationally equivalent. 
$\blacktriangleright$ The vertices not on a path from the input vertex to the output vertex do not contribute to the computation, so such vertices being pruned will not result in a different architecture cell. 
Considering the isomorphic graphs, NAS-Bench-101 consists of $\sim423$K unique architectures. For simplicity, the hard embeddings and soft embeddings of operation features are the same as NAS-Bench-201, we ignore the detailed description here. To be noticed, the nature of `validity constraints' and `isomorphic graphs' brings up challenges to our independent and unconstrained optimization of topological structure and operation features. We address this issue by deduplication, validity verification, and repeated sampling attempts. Interested readers are advised to refer to our implementation code.
\subsubsection{NAS-Bench-301.}
\begin{figure}[!ht]
  \centering
  \subfloat[][\label{fig:301-graph-fm-0}]{
    \includegraphics[width=0.35\columnwidth]{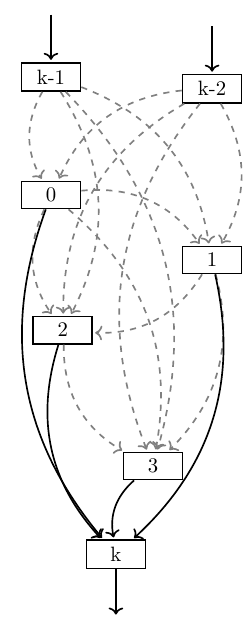}
  }
  \subfloat[][\label{fig:301-graph-fm-1-2}]{
    \includegraphics[width=0.45\columnwidth]{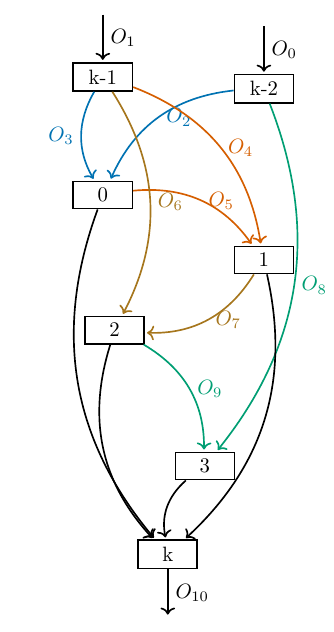}
  }
  \hfill
  \subfloat[][\label{fig:301-graph-node}]{
    \includegraphics[width=0.8\columnwidth]{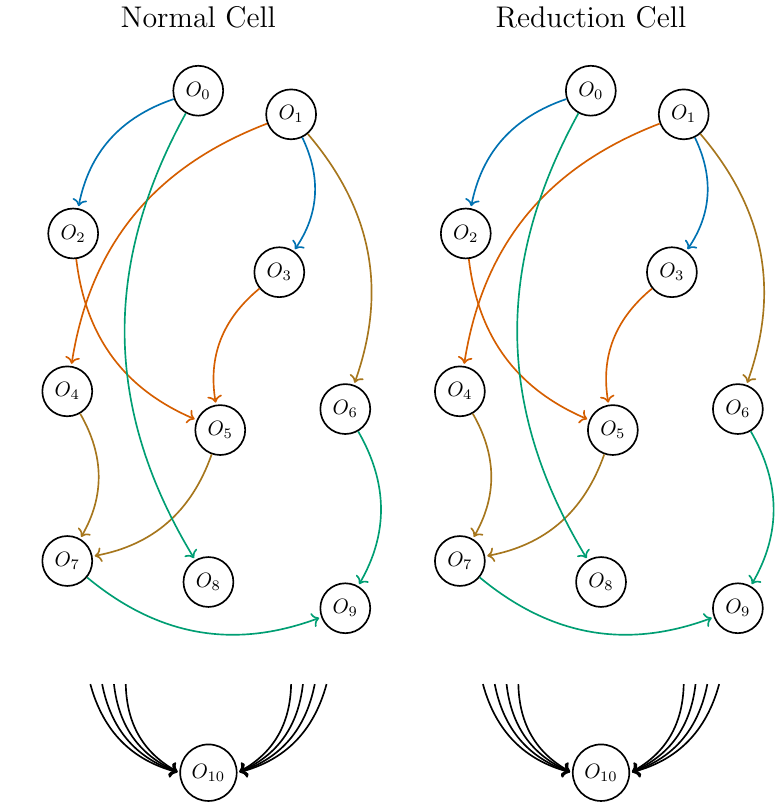}
  }
  \caption{Illustration of the search space of NAS-Bench-301. (a) The original schematic of be searched architecture cell, in the view of feature maps as the nodes and operations as the edge. The topological structure (edges represented by dashed line) and operation features (nodes) are to be searched, both for ``Normal cell'' and ``Reduction cell''; (b) The sampled architecture cell with $2$ proceeding feature maps (nodes); (c) The transformed graph representation with nodes representing operations and edges standing for the feature maps.}
  \label{fig:301-graph}
\end{figure}
\begin{figure}[!ht]
  \centering
      \includegraphics[width=0.7\columnwidth]{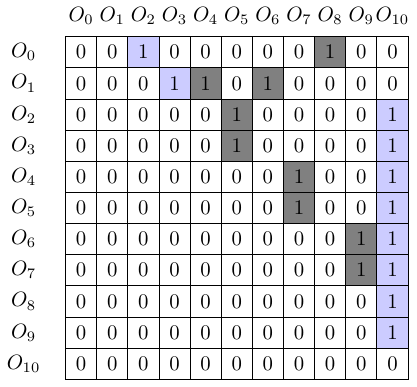}
  \caption{The illustration of the adjacency matrix of the sampled architecture. The elements with the blue background represent fixed edges, while elements with the lightgray background represent edge choices through sampling.}
  \label{fig:301-adj}
\end{figure}

\begin{figure}[!ht]
  \centering
    \subfloat[The relaxed topological structure variables $\beta$;\label{fig:301-proxy-a}]{
      \includegraphics[width=0.7\columnwidth]{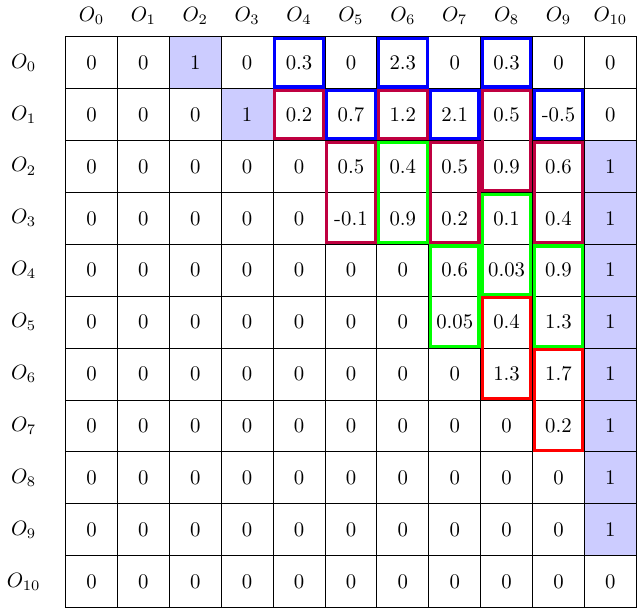}
    }
    \hspace{0.02\textwidth}
    \subfloat[The soft embedding of $\hat{s}_t$ based on the propagation from $\beta$, based on a simulative calculation with a temperature of $0.7$;\label{fig:301-proxy-b}]{
      \includegraphics[width=0.7\columnwidth]{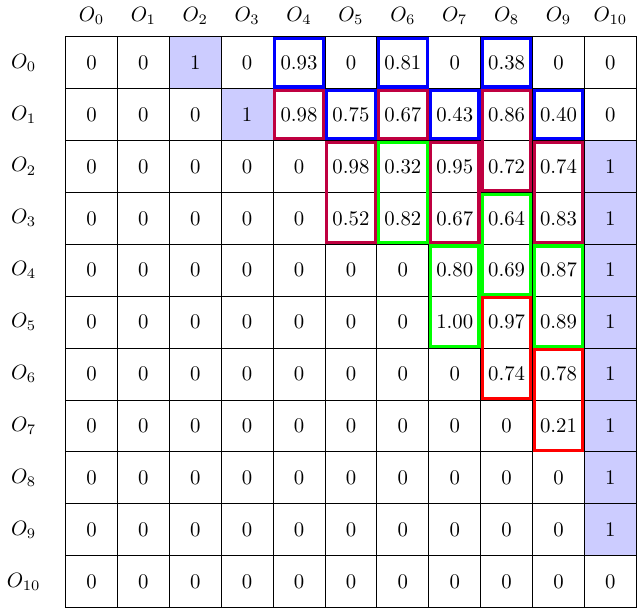}
    }
  \caption{The proxy sampling of topological structure variables in NAS-Bench-301. The colored boxes represent edge groups transformed based on the proceeding nodes relationship.}
  \label{fig:301-proxy-sampling}
\end{figure}
\begin{figure}[!ht]
  \centering
    \subfloat[The adjacency matrix with the mean value of group-wise elements after $\sigma(\beta)$ propagation;\label{fig:301-sampling-a}]{
      \includegraphics[width=0.7\columnwidth]{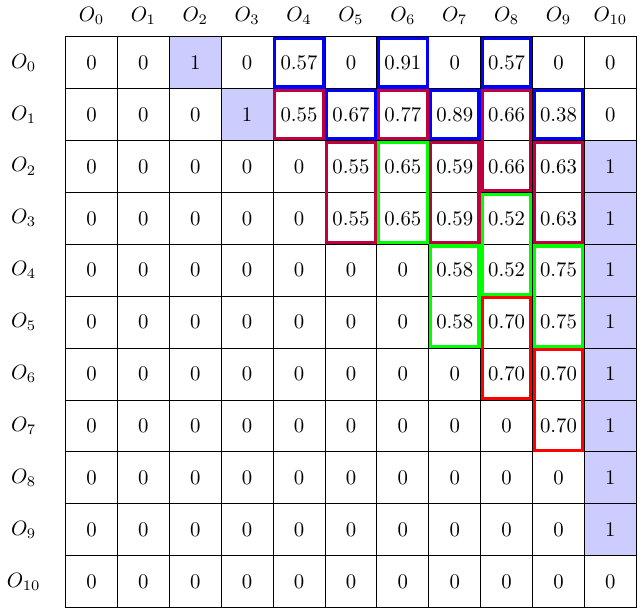}
    }
    \hspace{0.02\textwidth}
    \subfloat[The hard embedding with sampling ${s}_t$ function following the $\sigma(\beta)$;\label{fig:301-sampling-b}]{
      \includegraphics[width=0.7\columnwidth]{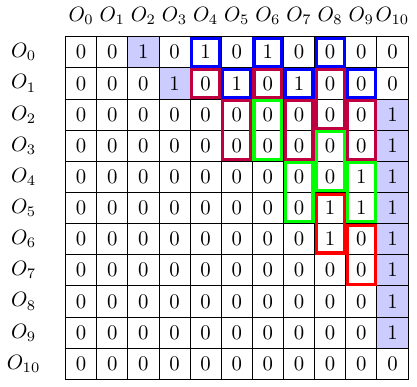}
    }
  \caption{The discrete sampling of topological structure variables in NAS-Bench-301.}
  \label{fig:301-sampling}
\end{figure}
\begin{table}[!t]
\centering
\caption{Candidate Preceding Nodes (Groups)}
\resizebox{0.45\textwidth}{!}{%
\begin{tabular}{@{}clc@{}}
\toprule
Node & Candidate Preceding Nodes (Groups) & Selection num \\ \midrule
$O_2$   & \{$O_0$\}                    & 1       \\
$O_3$   & \{$O_1$\}                    & 1       \\
$O_4$   & \{\{$O_0$\}, \{$O_1$\}\}        & 1       \\
$O_5$   & \{\{$O_1$\}, \{$O_2$, $O_3$\}\}    & 1       \\
$O_6$   & \{\{$O_0$\}, \{$O_1$\}, \{$O_2$, $O_3$\}\} & 1   \\
$O_7$   & \{\{$O_1$\}, \{$O_2$, $O_3$\}, \{$O_4$, $O_5$\}\} & 1 \\
$O_8$   & \{\{$O_0$\}, \{$O_1$\}, \{$O_2$, $O_3$\}, \{$O_4$, $O_5$\}\} & 1 \\
$O_9$   & \{\{$O_1$\}, \{$O_2$, $O_3$\}, \{$O_4$, $O_5$\}, \{$O_6$, $O_7$\}\} & 1 \\
\bottomrule
\end{tabular}
}
\label{tab:preceding}
\end{table}
\begin{figure}[!ht]
  \centering
      \includegraphics[width=0.75\columnwidth]{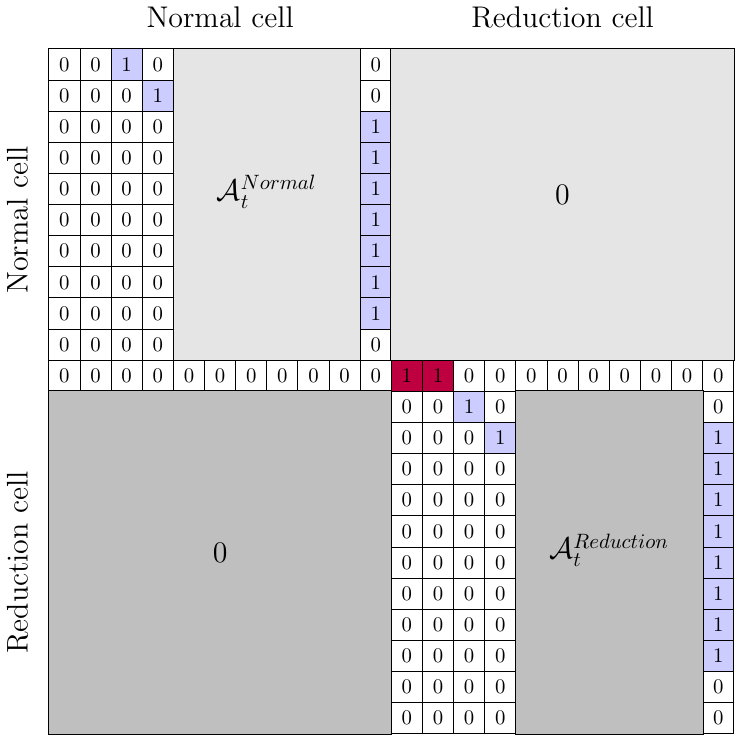}
  \caption{The combination of adjacency matrices of ``Normal cell'' and ``Reduction cell'', as an integrated graph structure.}
  \label{fig:301-adj-c}
\end{figure}
The candidate operation set of NAS-Bench-101 space is as follows:
\begin{minted}{Python}
opname_to_index = {
    'skip_connect': 0, 
    'max_pool_3x3': 1, 
    'avg_pool_3x3': 2, 
    'sep_conv_3x3': 3,
    'sep_conv_5x5': 4, 
    'dil_conv_3x3': 5, 
    'dil_conv_5x5': 6
}
\end{minted}
The space of NAS-Bench-301 is vast, containing $\sim10^{18}$ architectures, many orders of magnitude larger than any previous tabular NAS benchmark. For the scale of search space, surrogate models (\texttt{XGBoost, GIN and LGBoost}) are employed to make it as surrogate benchmarks. To note, for the types and structure of proxy models (neural networks), we employ the \texttt{XGBoost} as the surrogate model for experiments, keeping compliant with comparison methods. We present the search space illustration in \textcolor{blue}{\pref{fig:301-adj}}. The original schematic of the searched architecture cell is presented as the directed acyclic graph (DAG), in the view of feature maps as the nodes and operations as the edge. The design of NAS-Bench-301 allows a node to select two of its preceding nodes as its input nodes, as in\textcolor{blue}{~\pref{fig:301-graph-fm-1-2}}. To keep the compatible encoding scheme and facilitate the search, we transform the representation of the graph so that nodes represent operations and edges stand for the feature maps, as in \textcolor{blue}{~\pref{fig:301-graph-node}}. First, the precedence relationships of the feature maps need to be converted into the precedence relationships of the operation nodes, as in\textcolor{blue}{~\pref{tab:preceding}}, commonly, one of the proceeding nodes group is to be selected as the input node. Thus, given this transformation, we need to reconstruct the implementation logic of the original discrete sampling function w.r.t the topological structure, as well as the forward propagation logic of the proxy sampling propagation. Given the relaxed topological structure variables $\beta$ in \textcolor{blue}{~\pref{fig:301-proxy-a}}, which is constructed based on \textcolor{blue}{~\pref{tab:preceding}}: $\blacktriangleright$ For proxy sampling propagation, the soft embedding of $\hat{s}_t$ propagation based on $\beta$, as\textcolor{blue}{~\pref{fig:301-proxy-b}}. $\blacktriangleright$ For the discrete sampling function, the sigmoid propagation is element-wise, followed by the mean operation of group-wise elements, as \textcolor{blue}{~\pref{fig:301-sampling-a}}. Then the discrete sampling is based on the mean value of the groups, as\textcolor{blue}{~\pref{fig:301-sampling-b}}. $\blacktriangleright$ In terms of the chain-style structure with ``Normal cell'' and ``Reduction cell'' in NAS-Bench-301, we merge the two adjacency matrices as\textcolor{blue}{~\pref{fig:301-adj-c}}, combined with the concatenation of two operation feature matrices, as a fusion graph for propagation. 

Based on our practice, this transform strategy and encoding scheme of NAS-Bench-301 is meticulously designed but also reliable, reasonable, and effective. We recommend that readers refer to our code implementation. In addition, for simplicity, the hard embedding and soft embedding of operation features are the same as NAS-Bench-201 and NAS-Bench-101, we ignore the detailed description here. 
\subsubsection{NAS-Bench-NLP.}
NAS-Bench-NLP benchmark results from the recurrent neural network space. We keep the graph structure encoding scheme for NLP in the NASLib library unchanged, and do not specially design the discrete sampling function and proxy sampling propagation. Given the detailed descriptions of the aforementioned search spaces, we do not elaborate further here and recommend interested readers to examine the code.

\subsection{H. Hyperparameter and Reproducibility}
\label{sec:repreoduce}
In this section, we provide the hyperparameter settings and tuning experiences in our experiments. Overall, they are divided into two parts: \texttt{Model Fitting} and \texttt{Search and sampling}. For the same search space, the fitting parameters of the surrogate model generally remain consistent. However, for different search spaces, it is necessary to adjust the scale of the surrogate model based on the complexity of the space's graph structure, as well as the encoding schemes. Importantly, search and sampling parameters need to be empirically adjusted according to the space scale, query cost budget, and other factors. When dealing with real NAS tasks, hyperparameters can be adjusted using surrogate tasks or proxy metrics, e.g., early stop training criteria, or small-scale proxy task datasets. For instance, when transferring the hyperparameters setting for CIFAR-10 to CIFAR-100 and ImageNet16-120, the search results are still satisfactory.
\subsubsection{Hyperparameter setup and tuning experience} 
\label{sec:hypertuning}
\begin{table}[!ht]
\centering
\caption{Model fitting hyperparameters}
\label{tab:model_hyperparameters}
\setlength{\tabcolsep}{1mm}
\resizebox{0.45\textwidth}{!}{
\begin{tabular}{lll}
\toprule
\textbf{Parameter} & \textbf{Value} & \textbf{Description} \\ 
\midrule
\textit{gcn\_hidden}             & $144$            &  No. of neurons in a GCN layer\\
\textit{gcn\_layers}             & $2$              &  No. of GCN layers\\
\textit{linear\_size}             & $144$            &  No. of neurons in the MLP\\
\textit{batch\_size}             & $7$              &  Batch size\\
\textit{optimizer}             & AdamW              &  Optimizer for training\\
\textit{lr}                      & $0.001$          &  Learning rate\\
\textit{model\_epochs}           & $100$            &  No. of training epochs\\
\textit{ranking\_coe}  & $0.2$            &  Coefficient for ranking loss\\
\textit{m}               & $0.1$       & Ranking loss margin\\
\textit{lr\_scheduler}       & exponential    &  Learning rate scheduler\\
\textit{lr\_gamma}               & $0.9$            &  Decay rate of the lr scheduler\\
\textit{lr\_step}                & $10$             &  Step size of learning rate decay \\
\bottomrule
\end{tabular}
}
\end{table}
\begin{table}[!ht]
\centering
\caption{Search and sampling hyperparameters}
\label{tab:search_hyperparameters}
\setlength{\tabcolsep}{1mm}
\resizebox{0.45\textwidth}{!}{
\begin{tabular}{lll}
\toprule
\textbf{Parameter} & \textbf{Value} & \textbf{Description}  \\ 
\midrule
\textit{base\_temp}         & $0.7$  & Highest temperature\\
\textit{min\_temp}          & $0.2$  & Lowest temperature\\
\textit{lr\_alpha}          & $0.02$ & Learning rate for $\alpha$\\
\textit{lr\_beta}            & $0.001$ & Learning rate for $\beta$ \\
\textit{search\_epochs}            & $300$  & Optimizing epochs of each step \\
\textit{o\_epochs}            & $15$  & Optimization interval epochs for $\alpha$ \\
\textit{t\_epochs}            & $15$  & Optimization interval epochs for $\beta$ \\
\textit{optimizer1}   &  AdamW &  Optimizer for $\alpha$ \\
\textit{optimizer2}   &  AdamW &  Optimizer for $\beta$ \\
\textit{parallel\_batch}        & $5$    & $(\alpha, \beta)$ groups No. ($K$) \\
\textit{num\_sample}   & $10$   & Samples No. of each step \\
\textit{num\_sample\_init}         & $10$   & Initially random samples No.   \\
\textit{search\_steps}             & $9$    & No. of search steps \\
\textit{lhs\_lower}         & $0.9$  & The lower bound for LHS\\
\textit{lhs\_range}         & $0.1$  & The range bound for LHS\\
\textit{verify\_ratio}             & $20$   & Selection ratio ($Q/K$)\\
\bottomrule
\end{tabular}
}
\end{table}
We present the hyperparameters example of model fitting in \textcolor{blue}{\pref{tab:model_hyperparameters}}. Accordingly, we provide the following hyperparameter tuning insights:
    $\blacktriangleright$ The number of GCN layers affects the range of message passing, a significant increase in the GCN layers will cause over-smoothing. Under our framework, across all the NAS tasks, two layers are sufficient.
    $\blacktriangleright$ Typically, a reasonably bigger \textit{batch\_size} is better. If you aim to evaluate $100$ architectures in total, set \textit{batch\_size} to $20$ is adequate. But a large \textit{batch\_size} complicates ranking loss calculations, increasing the time consumption;
    $\blacktriangleright$ As for \textit{model\_epochs}, according to our practice, for common search spaces, $100$ training epochs are sufficient for prediction performance.
    $\blacktriangleright$ Setting the \textit{ranking\_coe} too high may bias the model predictions, while setting it too low may fail to preserve the prediction ranking performance.\\
We present the hyperparameters example of architecture search (gradient-based optimization) and sampling in \textcolor{blue}{\pref{tab:search_hyperparameters}}. Accordingly, we provide the following hyperparameter tuning insights: 
    $\blacktriangleright$ The temperature in Gumbel-Softmax can influence the randomness of the sampling. The predefined highest \textit{base\_temp} affects the exploration in the early stages of the search, within a range of $[0,\infty]$. Based on our experience, it is commonly set to $[0.5, 1]$, which is considered appropriate. The \textit{min\_temp} determines the unbiased nature of the proxy gradient and the original gradients, commonly set close to $0$.
    $\blacktriangleright$ The learning rate for $\alpha$ and $\beta$, their decay rate, and the Gumbel temperature must be empirically co-adjusted. If a particular operation or topological edge selection dominates from the outset, exploration should be encouraged, for example, by lowering the learning rate and increasing the range of temperature values.
    $\blacktriangleright$ The \textit{parallel\_batch} controls the group number of $(\alpha,\beta)$. Increasing it encourages exploration, but setting it significantly higher than \textit{num\_sample} is forbidden. \textit{num\_sample\_init} controls the number of points randomly sampled in the first search step. A large \textit{num\_sample\_init} can avoid local minima but may reduce efficiency. 
    $\blacktriangleright$ The \textit{verify\_ratio} determines the number of samples following $(\alpha,\beta)$, that is the selection ratio $Q/K$. 
    $\blacktriangleright$ During the initialization of the alphas, we use LHS to initialize them, with \textit{lhs\_range} controlling their separation.

\subsubsection{Environment Details and Reproducibility}
Our hardware and software specs are:
\begin{minted}{Python}
CPU: 13th Gen Intel Core i7-13700K
RAM: 128GB DDR5@5200MT/s
GPU: NVIDIA RTX 4090@2775MHz
Kernel: Linux 6.2.0-31-generic
NVIDIA Driver Version: 535.104.05
NVML Version: 12.535.104.05
CUDA Version: 11.8
PyTorch Version: 2.0.0
\end{minted}
Our experimental code is based on the repo of \url{https://github.com/automl/NASLib}, which involves the search space, dataset settings, original discrete encoding schemes, and data preprocessing from this excellent open-source library. This also facilitates our subsequent open-source contributions and future community engagement.
The code will be released at \url{https://github.com/blyucs/OptiProxy-NAS}. Python environment details are provided in a requirement file. If you have difficulty reproducing our work, please don't hesitate to contact us.




\end{document}